%% file: main.tex
\documentclass[letterpaper, 10 pt, journal, twoside]{ieeetran}
\IEEEoverridecommandlockouts
\let\labelindent\relax

\input{math_commands.tex}

\usepackage{times}
\usepackage{amsmath,amssymb,amsfonts}
\usepackage{mathtools}
\usepackage{multicol}

\usepackage[shortlabels]{enumitem}
\usepackage{float}
\usepackage{graphicx}
\usepackage{xcolor}
\usepackage{siunitx}
\usepackage[font=footnotesize,skip=0pt]{caption}
\usepackage[colorlinks,bookmarksopen,bookmarksnumbered,citecolor=black,urlcolor=black]{hyperref}
\hypersetup
{
	pdftitle = { Multi-Contact Inertial Parameters Estimation and Localization in Legged Robots},
	pdfauthor = {Sergi Martinez, Robert J. Griffin, Carlos Mastalli},
	pdfsubject = {Submitted to IEEE Robotics and Automation Letters (RA-L)},
	pdftoolbar = true,
	colorlinks = true,
	linkcolor = black,
	citecolor = black,
	urlcolor = black,
}

\usepackage{glossaries}
\usepackage{multirow}
\usepackage{hhline}
\usepackage{booktabs}
\usepackage{tikz}
\newcommand{\tikzxmark}{%
\tikz[scale=0.23] {
    \draw[line width=0.7,line cap=round] (0,0) to [bend left=6] (1,1);
    \draw[line width=0.7,line cap=round] (0.2,0.95) to [bend right=3] (0.8,0.05);
}}
\newcommand{\ra}[1]{\renewcommand{\arraystretch}{#1}}
\glsdisablehyper
\newacronym{oe}{OE}{optimal estimation}
\newacronym{nlp}{NLP}{nonlinear programming}
\newacronym{kkt}{KKT}{Karush-Kuhn-Tucker}
\newacronym{ddp}{DDP}{differential dynamic programming}
\newacronym{mpc}{MPC}{model predictive control}
\newacronym{aba}{ABA}{articulated body algorithm}
\newacronym{imm}{IMM}{inertial matrix method}
\newacronym{id}{ID}{inverse dynamics}
\newacronym{fd}{FD}{forward dynamics}
\newacronym{ekf}{EKF}{extended Kalman filter}

\usepackage{cleveref}
\Crefname{definition}{Definition}{Definitions}
\Crefname{proposition}{Proposition}{Propositions}
\Crefname{theorem}{Theorem}{Theorems}
\Crefname{figure}{Fig.}{Figs.}
\Crefname{equation}{Eq.}{Eqs.}
\Crefname{section}{Section}{Sections}
\Crefname{subsection}{Section}{Sections}
\Crefname{subsubsection}{Section}{Sections}
\Crefname{algorithm}{Algorithm}{Algorithms}
\newcommand{\video}{https://youtu.be/nDOxTgeK_5o}

\title{Multi-Contact Inertial \REV{Parameters} Estimation and Localization in Legged Robots}
\author{Sergi Martinez\textsuperscript{\normalfont 1}\quad
Robert \finalrev{J.} Griffin\textsuperscript{\normalfont 2}\quad
Carlos Mastalli\textsuperscript{\normalfont 1,2}
\thanks{\textsuperscript{\normalfont 1} Robot Motor Intelligence (RoMI) Lab -- Heriot-Watt University, UK.}
\thanks{\textsuperscript{\normalfont 2} IHMC Robotics -- Florida Institute for Human \& Machine Cognition, US.}
}

\begin{document}

\maketitle

\begin{abstract}
\rev{Optimal estimation is a promising tool for estimation of payloads' inertial parameters and localization of robots in the presence of multiple contacts.}
To harness its advantages in robotics, it is crucial to solve these large and challenging optimization problems efficiently.
\rev{To tackle this, we (i) develop a multiple shooting solver that exploits both temporal and parametric structures through a parametrized Riccati recursion.
Additionally, we (ii) propose an \textit{inertial manifold} that ensures the full physical consistency of inertial parameters and enhances convergence}.
To handle its \rev{manifold} singularities, we (iii) introduce a nullspace approach in our optimal estimation solver.
\rev{Finally, we} (iv) develop the analytical derivatives of contact dynamics for both inertial parametrizations.
Our framework can successfully solve estimation problems for complex maneuvers such as brachiation in humanoids, \REV{achieving higher accuracy than conventional least squares approaches}.
We demonstrate its numerical capabilities across various robotics tasks and its benefits in experimental trials with the Go1 robot.
\end{abstract}

\IEEEpeerreviewmaketitle

\section{Introduction}
\begin{figure}[t]\centering
    \href{\video&t=28}{\includegraphics[width=0.95\linewidth]{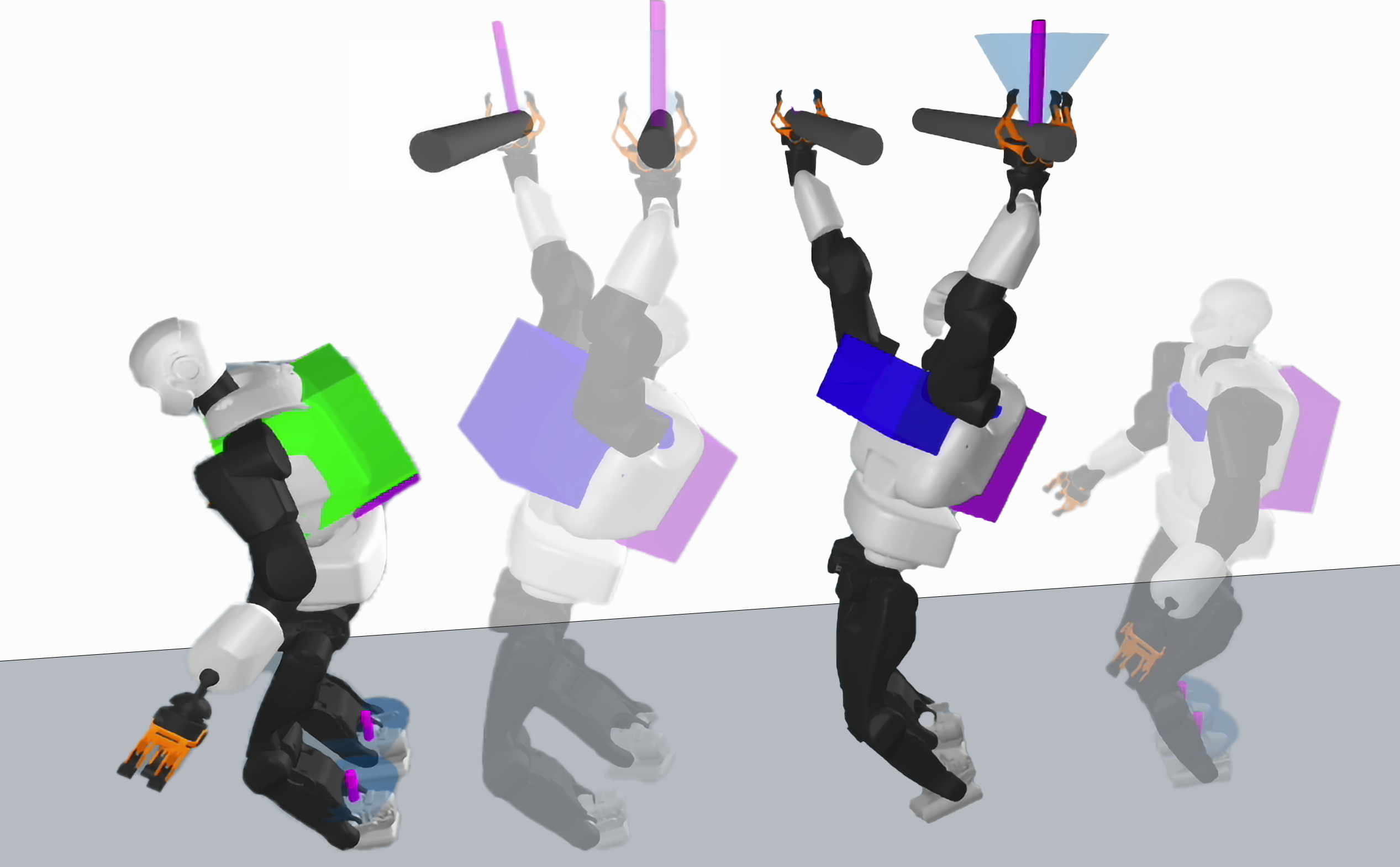}}
    \caption{Talos executing intricate monkey bar maneuvers with an unidentified payload.
    In the initial stages, our estimator meticulously pinpoints Talos' localization and estimates its payload's properties: mass, barycenter, and rotational inertia.
    The \rev{boxes' sizes}, positions, \rev{shape}, and colors serve as visual indicators, representing estimations of the payload's mass, barycenter, principal components of inertia, and algorithm convergence.
    To watch the video, click the picture or see \texttt{\url{\video}}.}
    \label{fig:cover}
\end{figure}

\IEEEPARstart{O}{ptimal estimation}~(\acrshort{oe})~\cite{optest2004textbook} emerges as a powerful tool for interpreting observations and accurately estimating a system's true state, including internal changes like unknown payloads.
In the context of robotics, this framework takes into account both proprioceptive and exteroceptive observations~\cite{foster-tro17,wish-ral23}.
It systematically incorporates considerations such as robot dynamics and nonholonomics~\cite{wieber-fmbr05}, balance conditions~\cite{orsolino-ral18}, and kinematic range.
Its application in robotics holds promising potentials.
\rev{For instance, it can be applied to estimate external parameters, e.g., aerodynamics effects in aerial navigation or internal parameters, e.g., unknown payloads as shown in \Cref{fig:cover}.}

Algorithms for optimal estimation solve variations of the following problem:
\begin{align}\label{eq:oe_problem}\nonumber
\min_{\stateSeq, \ucertainSeq,\params}
&\,\, \frac{1}{2}\|\state[0]\ominus\stateMean[0]\|^2_{\stateCov[0]^{-1}} + \frac{1}{2}\|\params-\paramsMean\|^2_{\paramsCov^{-1}}\\\nonumber
&\hspace{-0.5em}+\frac{1}{2}\sum_{k=0}^{N-1}\|\ucertain[k]\|^2_{\ucertainCov[N]^{-1}}+\frac{1}{2}\sum_{j=1}^{N}\|\obsMeas[j]\ominus\obsFunc(\state[j];\params\vert\ctrlMeas[j])\|^2_{\obsCov[j]^{-1}}\\
&\text{s.t.} \quad \state[k+1]=\dynFunc(\mathbf{x}_{k};\params\vert\ctrlMeas[k])\oplus\ucertain[k], \hspace{1em}\state[N] = \state[N-1],
\end{align}
where $\state\in\stateManif\subseteq\R^\nx$ represents the system's state, $\ucertain\in\stateTManif\subseteq\R^\nx$ describes its uncertainty, $\params\in\R^\nparams$ are static parameters defining the system's inertial properties, $\mathbf{\hat{u}}_{k}$ denotes the applied (and known) control commands \REV{(e.g., joint torques)}, $\obsMeas\in\obsManif\subseteq\R^\nz$ are observations, and $\state[0]$ is the arrival state.
\Cref{eq:oe_problem} \rev{finds} the \textit{maximum a-posteriori estimate} $P(\stateSeq,\ucertainSeq,\params\vert\obsMeas[s])$ as the state, parameters, uncertainties, and observations correspond to Gaussian distributions, i.e., $\state[k]\thicksim\mathcal{N}(\stateMean[k],\stateCov[k])$, $\params\thicksim\mathcal{N}(\paramsMean,\paramsCov)$, $\ucertain[k]\thicksim\mathcal{N}(\mathbf{0},\ucertainCov[k])$, and $\obsMeas[k]\thicksim\mathcal{N}(\mathbf{0},\obsCov[k])$, respectively.
The term $\frac{1}{2}\|\state[0]\ominus\stateMean[0]\|^2_{\stateCov[0]^{-1}}$ describes the \textit{arrival \rev{state distribution}}, where $\ominus$ is the \textit{difference} operator used to optimize over state manifolds~\cite{gabay82jota,mastalli-icra20,sola2018micro}.
Additionally, the term $\frac{1}{2}\|\params-\paramsMean\|^2_{\paramsCov^{-1}}$ specifies the uncertainties in the system's parameters.
Finally, $\dynFunc:\stateManif\times\R^\nparams\to\stateTManif$ and $\obsFunc:\stateManif\times\R^\nparams\to\obsManif$ are nonlinear functions describing the \rev{robot} dynamics and measurements.
\rev{The observation set $\obsMeas\in\obsManif$ depends on the onboard sensors available on a robot.
For example, in legged robots, this typically includes $\obsMeas=[\posMeas[j]\,\, \velMeas[j]\,\, \accMeas[i]\,\, \angVelMeas[i] \,\, \forcelMeas[c]]^\transpose$, where $\posMeas[j]$ and $\velMeas[j]$ are the joint positions and velocities, $\accMeas\in\R^{3}$ and $\angVelMeas\in\R^{3}$ are the IMU linear accelerations and angular velocities, and $\forcelMeas[c]$ are the contact forces used to define the contact constraints.} 

\subsection{Related work}
To address optimal estimation in robotics, one can leverage well-established \textit{direct methods}~\cite{betts-bookoptctrl}, which transcribes~\Cref{eq:oe_problem} into a~\gls{nlp} problem.
Direct methods involve the discretization of both state and uncertainties, followed by optimization using sparse general-purpose \gls{nlp} software such as \textsc{Snopt}~\cite{gill-siam05}, \textsc{Knitro}~\cite{byrd-knitro06}, and \textsc{Ipopt}~\cite{wachter-mp06}.
These software rely on sparse linear solvers such as MA27, MA57, and MA97 (see~\cite{HSL}) to factorize the large~\gls{kkt} problem.
However, a limitation of these linear solvers is their inefficiency in exploiting the Markovian structure of~\gls{oe} problems~\cite{mastalli22auro}, restricting their applicability in real-time applications, especially in legged robotics.
These computational limitations are attributed to their difficulties in utilizing data cache accesses efficiently, resulting in the exclusion of~\gls{oe} strategies.
Indeed, recent works focused on factor graphs formulations \REV{with robot kinematics}~\cite{dellaert2017factorgraphs,harley2018contactfactors,agrawal22factorgraph} are restricted to localization approaches only, ignoring the robot's dynamics.
In contrast, our approach considers the robot's dynamics, thereby reducing drift errors in \textit{proprioceptive localization} (\rev{see~\Cref{fig:estimationGO1_pos}-bottom}).

\begin{figure*}[t]
\href{\video&t=10}{\includegraphics[width = 1.0\linewidth]{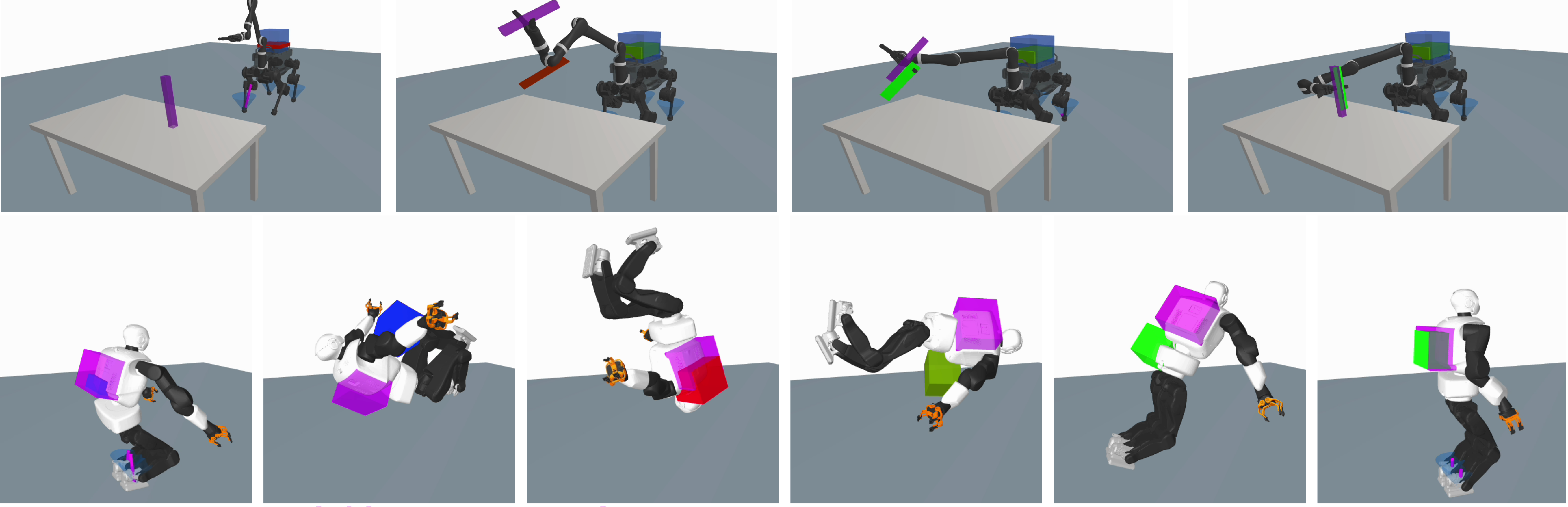}}
\caption{Snapshots illustrating  simulated online multi-contact inertial estimation and localization.
(top) ANYmal with a Kinova arm carries an unknown payload while grasping an unknown object.
(\REV{bottom}) Talos performs a backflip while simultaneously estimating its backpack.
Unknown payloads are represented in magenta, and upon convergence of the estimator, the estimated payload is depicted in green. \rev{The horizon of the estimator is set to the length of the simulation.}
To watch the video, click the picture or see \texttt{\url{\video}}.}
\label{fig:snapshots}
\end{figure*}

Alternatively, we can leverage Bellman's principle of optimality~\cite{bellman54bull} to break the optimal estimation problem into a sequence of smaller subproblems.
This approach effectively exploits the Markovian structure, resulting in a \gls{ddp} approach for optimal estimation~\cite{kobilarov2015optestddp}.
Additionally, by augmenting the system's state with its internal parameters, or parametrizing the dynamics, we can simultaneously solve identification and localization problems via~\gls{ddp}~\cite{oshin2022pddp}.
However, these \gls{ddp} approaches, being \textit{single shooting} algorithms, are prone to numerical instability and require a good initialization~\cite{betts-bookoptctrl}, both crucial considerations for their application in robotics.
These numerical instabilities, arise from enforcing dynamic feasibility, can be mitigated through \textit{feasibility-driven} methods~\cite{mastalli22auro,mastalli-invdynmpc} or \textit{multiple-shooting} strategies~\cite{li2023multshootddp} as proposed in~\gls{mpc} literature.
Moreover, \gls{ddp} approaches for optimal estimation presuppose knowledge of the \textit{arrival state}, a condition not attainable in real-world applications.
In contrast, our framework incorporates the \rev{\textit{arrival state estimation}}.

To estimate inertial parameters in robotics, two key aspects must be leveraged: (i) the affine relationship between these parameters and the generalized torques, as initially developed in~\cite{atkeson1986estimation}, and (ii) the analytical derivatives of rigid body algorithms~\cite{carpentier2018analytical,singh2022svaderivative}.
The second aspect is particularly relevant to us, as the derivatives of forward dynamics rely on the~\gls{imm} outlined in~\cite{featherstone2014rigid}.
This is because \Glspl{imm} involve Cholesky decompositions of the joint-space inertia matrix, limiting their operation to inertial parameters that are \textit{fully physically consistent}.

Conditions for full physical consistency boils down to triangle inequality constraints~\cite{traversaroidentification}.
However, numerical optimization guarantees inequality feasibilities at convergence, limiting its application to~\glspl{imm}.
Alternatively, these conditions can be embedded in a smooth manifold defined via a log-Cholesky parametrization~\cite{rucker2022smooth}.
This manifold, being singularity-free, exhibits highly nonlinear geometry.
\REV{Alternatively}, it is possible to build a smoother manifold by developing an~\gls{oe} solver that handles these singularities.
These ideas lead us to a novel \textit{exponential eigenvalue manifold}, with a better convergence rate, coupled with a nullspace resolution (\rev{see~\Cref{fig:parametrizations}}).

\subsection{Contribution}
Our main contribution is an efficient approach for solving hybrid optimal estimation problems in \rev{inertial identification} and localization.
It relies on \rev{three} technical contributions:
\begin{enumerate}[(i)]
    \item a novel smooth manifold that guarantees full physical consistency of inertial parameters \rev{(\Cref{sec:expeig_parametrization})},
    \item analytical derivatives of the hybrid contact dynamics regarding the inertial parameters \rev{(\Cref{sec:OE})}, and
    \rev{\item a novel optimal estimation solver that handles singularities in inertial identification, arrival state estimation, and incorporates multiple shooting rollouts (\Cref{sec:PDDP}).}
\end{enumerate}

Our optimal estimation framework is the first of its \REV{kind and integrates the state-of-the-art} log-Cholesky parametrization.
This is because \REV{current frameworks (e.g.,}~\cite{rucker2022smooth}) focuses solely on inertial identification \REV{via a \textit{maximum likelihood approach}}.
Additionally, \REV{we develop} a novel multiple shooting solver that combines nullspace parametrization to handle the exponential eigenvalue singularities.
Such singularities occur when the principal components of inertia at barycenter are the same, as in a solid sphere or a disk with uniform density.

\rev{The article is structured as follows: \Cref{sec:background} introduces contact dynamics, inertial parameters, and their conditions to physical consistency.
To satisfy these conditions, \Cref{sec:parametrization} derives our exponential eigenvalue parametrization. \REV{\Cref{sec:OE} describes the optimal estimation problem.}
\Cref{sec:PDDP} covers technical details of our optimal estimation solver: the parametrized Riccati recursion, multiple shooting rollouts, and arrival state estimation.
\Cref{sec:reuslts} \REV{presents} results \REV{that demonstrate the benefits of our approach, while},~\Cref{sec:conclusion} provides the conclusion.
}

\section{Background}\label{sec:background}

\subsection{Contact dynamics}\label{sec:contact_dyn}
The dynamics of rigid body systems, subject to \rev{holonomic} contact constraints at the acceleration level, are governed by:
\begin{align}\label{eq:dynamics}
\massMatrix(\pos)\acc &= \eff(\mathbf{u}) - \coriolisGravTerm(\pos,\vel) + \contactJac(\pos)^\transpose\contactForce,\\\label{eq:contact_constraint}
\contactJac(\pos)\acc &= -\contactAcc(\pos,\vel),
\end{align}
where $\pos\in\confManif\subseteq\R^{\nq}$, $\vel\in\confTManif\subseteq\R^{\nq}$,  $\mathbf{u}\in\R^{\ntau}$, $\contactForce\in\R^\nc$ \rev{and $\contactAcc\in\R^\nc$} represent the configuration point, generalized velocity, control commands, contact forces, \rev{and contact acceleration}, respectively. 
The functions $\massMatrix:\confManif\to\R^{\nq\times\nq}$ represent the joint-space inertia matrix, $\eff:\R^\ntau\to\R^\nq$ denotes the joint-generalized torque, $\contactJac:\rev{\confManif}\to\R^{\nc\times\nq}$ is the contact Jacobian, and $\coriolisGravTerm:\confManif\times\confTManif\to\R^\nq$ is a force vector containing the Coriolis, centrifugal and gravitational terms.
In the absence of contacts, free dynamics are governed by \Cref{eq:dynamics} with $\contactForce=\mathbf{0}$. 
\rev{We define unilateral conditions of \Cref{eq:contact_constraint} by penalizing negative contact forces~\cite{mastalli2022agile}.
Finally, this contact model neglects slipagges and the presence of joint friction.}

Free dynamics are \REV{efficiently} computed using the~\gls{aba}~\cite{featherstone2014rigid}.
In contrast, contact dynamics are commonly computed using a Schur-complement approach based on the~\gls{imm}~\cite{budhiraja-ichr18}.
Moreover, analytical derivatives of contact dynamics with respect to $\pos$, $\vel$ and $\eff$ are computed based on the algorithms described in~\cite{carpentier2018analytical}, as explained in~\cite{mastalli-icra20}.
\REV{These algorithms for computing dynamics and their derivatives are essential for real-time applications.}

\subsection{Inertial parameters of rigid bodies}\label{sec:inertial_params}
The spatial inertia of a rigid body $i$, encapsulating the body's mass-density field $\massDensity:\R^3\to\R_{\geq 0}$, can be defined using a vector $\dynParams[i]\in\R^{10}$, whose elements are
\begin{equation}\label{eq:inertial_param_vector}
\setlength\arraycolsep{2pt}
\dynParams[i] = 
\begin{bmatrix}
\mass & h_x & h_y & h_z & I_{xx} & I_{xy} & I_{yy} & I_{xz} & I_{yz} & I_{zz}
\end{bmatrix}^\transpose,
\end{equation}
where $\mass\in\R$ denotes the body's mass, $\lever = [h_x\,\, h_y\,\, h_z]^\transpose = \mass\barycenter\in\R^3$ represents the first mass moment with $\barycenter$ as its barycenter, 
and $[I_{xx}\,\, I_{xy}\,\, I_{yy}\,\, I_{xz}\,\, I_{yz}\,\, I_{zz}]^\transpose$ are the elements of its inertia matrix $\rotInertia\in\R^{3\times 3}$.
Both barycenter and rotational inertia are expressed in the body-fixed reference frame as introduced in~\cite{atkeson1986estimation}.
Moreover, the rotational inertia $\rotInertia$ can be expressed at the barycenter using the parallel axis theorem:
\begin{gather}\label{eq:Ic}
\rotInertiaBarycenter = \rotInertia - \frac{
\skewMatrix(\mathbf{h})\skewMatrix(\mathbf{h})^\transpose}{m},
\end{gather}
where $\skewMatrix:\R^3\to\so[3]$ is the skew-symmetric matrix, also known as the Lie algebra of $\SO[3]$ (see~\cite{sola2018micro}).

As described in~\cite{atkeson1986estimation}, the generalized torques can be expressed as an affine function of the inertial parameters $\dynParams$, i.e.,
\begin{gather}\label{eq:NE_param} 
\eff(\ctrl) = \regMatrix(\pos,\vel,\acc)\dynParams,
\end{gather}
where $\regMatrix:\confManif\times\confTManif\times\confTManif\to\R^{\nq\times 10\nb}$ is the joint-torque regressor matrix with $\nb$ as the number of rigid bodies.

\subsection{Physically-consistent spatial inertia}
We say that $\dynParams[i]$ is \textit{fully physically consistent}, or $\dynParams[i]\in\inertialManif$, if there exists a finite-value density function $\massDensity$.
The set $\inertialManif\subset\R^{10}$ is characterized by the following inequality constraints:
\begin{gather}\label{eq:physical_consistent_cond}
\begin{aligned}
\mass \REV{>} 0, \quad &\rotInertiaBarycenter \succeq 0, \\
D_x < D_y + D_z, \quad D_y < D_x +& D_z, \quad D_z < D_x + D_y.
\end{aligned}
\end{gather}
Here, $\eigInertia=\text{diag}(D_x, D_y, D_z)\in\R^{3\times 3}$ represents the principal moments of inertia, computed as $\rotInertiaBarycenter=\rotMatrix\eigInertia\rotMatrix^\transpose$ with $\rotMatrix\in\SO[3]$.
The condition $\rotInertiaBarycenter \succeq 0$ ensures positivity in the principal components of inertia, and the triangle inequalities enforce positivity in second moments of mass~\cite{traversaroidentification}.

The rotational inertia $\rotInertiaBarycenter$ can be interpreted as a point-mass distribution with covariance $\covInertia[c]$.
When mapping this to its spatial inertia, it results in a 4x4 pseudo-inertia matrix~\cite{wensing2017linear}:
\begin{gather}\label{eq:pseudo_inertia}
\pseudoInertia = \begin{bmatrix}
\covInertia & \lever \\
\lever^\transpose & \mass
\end{bmatrix},
\quad \text{where} \quad \covInertia = \frac{1}{2}\Tr[\rotInertia]\eyeMatrix[3] - \rotInertia,
\end{gather}
$\covInertia = \covInertia[c] + \mass\barycenter\barycenter^\transpose$,  $\Tr[\cdot]$ is the trace operator, and $\eyeMatrix[3]$ is a 3x3 identity matrix.
The pseudo-inertia matrix $\pseudoInertia\in\pseudoInertiaManif[4]$ must satisfy the condition $\pseudoInertia\succeq 0$ for full physical consistency.

\section{Parametrization of inertial parameters}\label{sec:parametrization}
To map parameters $\params[\rev{i}]\in\R^{10}$ to fully physically consistent parameters $\dynParams[i]\in\inertialManif$, we can construct a smooth isomorphic mapping $\paramsMap[\rev{i}]:\R^{10}\to\inertialManif$.
Below, we \rev{present two isomorphic mappings}.
\rev{First, we introduce} log-Cholesky parametrization and then proceed to develop our \textit{exponential eigenvalue parametrization}.

\subsection{Log-Cholesky parametrization}
Log-Cholesky decompositions are applicable to semidefine matrices, enabling us to encode the condition $\pseudoInertia[i]\succeq 0$ for physical consistency of body $i$.
As suggested in~\cite{rucker2022smooth}, a log-Cholesky parametrization of the pseudo inertia $\pseudoInertia=\upperTriangMatrix\upperTriangMatrix^\transpose$, with\vspace{-0.8em}
\begin{gather}\label{eq:pseudo_inertia_logchole}
\upperTriangMatrix = e^\alpha
\begin{bmatrix}
e^{\dParam[1]} & \sParam[12] & \sParam[13] & \tParam[1] \\
0 & e^{\dParam[2]} & \sParam[23] & \tParam[2] \\
0 & 0 & \sParam[23] & \tParam[3] \\
0 & 0 & 0 & 1
\end{bmatrix},
\end{gather}
can be employed to establish an isomorphic map $\dynParams=\paramsMap(\params[\text{lchol}])$.
Here, $\paramsMap \coloneqq \text{vech}(\upperTriangMatrix\upperTriangMatrix^\transpose)$, where the $\text{vech}\REV{(\cdot)}$ operator
denotes the serialization operation of $\pseudoInertia$, and
\begin{gather}\label{eq:logchol_param}
\setlength\arraycolsep{2pt}
\params[\text{lchol}]=
\begin{bmatrix}
\alpha& \dParam[1]& \dParam[2]& \dParam[3]& \sParam[12]& \sParam[23]& \sParam[13]& \tParam[1]& \tParam[2]& \tParam[3]
\end{bmatrix}^\transpose \in \mathbb{R}^{10}.
\end{gather}

\subsection{Exponential eigenvalue parametrization}\label{sec:expeig_parametrization}
The singular value decomposition of $\rotInertiaBarycenter$ can be expressed in terms of its second moments of mass:\vspace{-0.5em}
\begin{gather}
\rotInertiaBarycenter = \rotMatrix
\,\overbrace{\text{diag}\left(\mathbf{P}\secMomentMass\right)}^\eigInertia\
\rotMatrix^\transpose, \,\,\text{where}
\,\,\mathbf{P}=\begin{bmatrix}
0 & 1 & 1\\
1 & 0 & 1\\
1 & 1 & 0
\end{bmatrix}
\end{gather}
and $\secMomentMass=[L_x\,\, L_y\,\, L_z]\in\R^3_{\geq 0}$ denotes the second moments of mass.
Ensuring positivity in $L_x$, $L_y$, $L_z$ is equivalent to satisfying both $\rotInertiaBarycenter\succeq 0$ and the triangle inequalities in~\Cref{eq:physical_consistent_cond}.
Intuitively, these conditions on $L_x$, $L_y$, $L_z$, and $\mass \geq 0$ can be embedded using an exponential map.
Formally, their manifold constraints are
\begin{gather}
\text{diag}(\secMomentMass)\cdot\text{diag}(\secMomentMass)^{-1}=\eyeMatrix[3], \quad \mass\cdot\mass^{-1} = 1,
\end{gather}
defining the real numbers as their Lie algebra structure.
Therefore, our parametrization is given by
\begin{gather}\label{eq:exp_eig}
\setlength\arraycolsep{2pt}
\mass = \text{exp}(\sigma_\mass), \quad \secMomentMass=
\begin{bmatrix}
\text{exp}(\sigma_x) & \text{exp}(\sigma_y) & \text{exp}(\sigma_z)
\end{bmatrix}^\transpose,
\end{gather}
where $\sigma_\mass$ and $(\sigma_x, \sigma_y, \sigma_z)$ denote the mass and rotational inertia parameters, respectively.
Moreover, the structure of the rotational matrix $\rotMatrix$ is guaranteed by parametrizing it via the Lie algebra of $\SO[3]$, i.e.,
\begin{gather}
\rotMatrix = \text{Exp}(\rotMatrixParam),
\end{gather}
where $\rotMatrixParam=[\omega_x\,\, \omega_y\,\, \omega_z]\in\R^3\cong\so[3]$, and $\text{Exp}:\R^3\to\SO[3]$ converts this vector elements to the elements of the $\SO[3]$ group.
Moreover, we obtain the inertia $\rotInertia$ via~\Cref{eq:Ic}.

Combining the exponential maps and stacking them into a vector, we construct a \rev{novel} map with local submersion $\dynParams=\paramsMap(\params[\text{eeval}])$, where
\begin{gather}\label{eq:eeval_param}
\setlength\arraycolsep{2pt}
\params[\text{eeval}]=
\begin{bmatrix}
\sigma_m& h_x& h_y& h_z& \omega_x& \omega_y& \omega_z& \sigma_x& \sigma_y& \sigma_z
\end{bmatrix}^\transpose \in \mathbb{R}^{10}.
\end{gather} 

Finally, we can analytically compute $\frac{\partial\dynParams}{\partial\params}$ for both $\params[\text{lchol}]$ and $\params[\text{eeval}]$, as they are smooth and differentiable parametrizations.

\section{Multi-contact optimal estimation}\label{sec:OE}
Optimal estimation with hybrid events involves the use of multiphase dynamics and reset maps:
\begin{align}\label{eq:oe_complete}
\min_{\stateSeq,\ucertainSeq,\params}
&\hspace{-2.em}
& & \hspace{-0.75em}\ell_N(\state[N];\params\vert\obsMeas[N])+\sum_{k=0}^{N-1} \ell_{k}(\state[k],\ucertain[k];\params\vert\obsMeas[k]) \hspace{-8.em}&\\\nonumber
&\hspace{-2.2em}\textrm{s.t.} & &\hspace{-1.8em}\text{for $p\in\mathcal{P}$:}\\\nonumber
&\hspace{-2.2em} & &\hspace{-0.4em}\text{for $k\in\{p_0, p_0 + 1, \cdots, p_N-1\}$:}\\\nonumber
& & &\hspace{0.8em} \state[k+1] = \dynFunc[p](\state[k];\params\vert\ctrlMeas[k])\oplus\ucertain[k] \hspace{-0.5em}&\hspace{2em}\textrm{(phase dyn.)}\\\nonumber
& & &\hspace{-0.4em} \state[p_{N+1}] = \resetFunc[p](\state[p_N];\params), \hspace{-1em}&\hspace{-1em} \,\,\, \textrm{(reset map)}
\end{align}
where $\dynFunc[p]:\stateManif\times\R^\nparams\to\stateTManif$ describes the \REV{free or} contact dynamics in phase $p$, 
$\resetFunc[p]:\stateManif\times\R^\nparams\to\stateTManif$ defines its contact-gain transition (modeled via impulse dynamics)\rev{, and $\ell_{k}$ encodes the observations $\obsMeas[k]$ at the $k$-th node}.
Below, we \rev{describe our novel} analytical derivatives w.r.t. $\params$, \rev{which is inspired from the} analytical derivatives w.r.t. $(\pos,\vel)$ \rev{described in}~\cite{mastalli-icra20,mastalli2022agile}.

\subsection{Analytical derivatives of parametrized dynamics}\label{sec:contact_derivatives}
In free dynamics, inertial parameters $\dynParams\rev{=(\dynParams[i])_{i=0}^{n_b}}$ \REV{(with $n_b$ denoting the number of bodies)} exhibit a linear relationship with the generalized torques \rev{and} $\partial\text{FD} = \massMatrix^{-1}\partial\text{ID}$ holds, \rev{as} the \gls{fd} is the reciprocal of~\gls{id}~\cite{carpentier2018analytical}.
Therefore, by applying the chain rule, we compute the analytical derivatives of forward dynamics with respect to $\params$ as
\begin{gather}\label{eq:fwddyn_derivaties}
\frac{\partial\text{FD}}{\partial\params} = \massMatrix(\pos)^{-1}\frac{\partial\text{ID}}{\partial\dynParams}\frac{\partial\dynParams}{\partial\params} = 
\massMatrix(\pos)^{-1}\regMatrix(\pos,\vel,\acc)\frac{\partial\dynParams}{\partial\params}.
\end{gather}

Moving on to contact\rev{/phase} dynamics \rev{$\dynFunc[p]$} \rev{of~\Cref{sec:contact_dyn}}, we apply the chain rule to derive the analytical derivatives for both the system's acceleration and the contact forces:
\begin{gather}\label{eq:diff_c_p}
\frac{\partial}{\partial\params}
\begin{bmatrix}
\acc \\ -\contactForce
\end{bmatrix} = -
\begin{bmatrix}
\massMatrix(\pos) & \contactJac(\pos)^\transpose \\
\contactJac(\pos) & \zeroMatrix
\end{bmatrix}^{-1}
\begin{bmatrix}
\frac{\partial\eff}{\partial\params} \\ \zeroVec
\end{bmatrix},
\end{gather}
where $\frac{\partial\eff}{\partial\params}=\regMatrix(\pos,\vel,\acc)\frac{\partial\dynParams}{\partial\params}$ \rev{and $\frac{\partial\dynParams}{\partial\params}$ is the partial derivatives of the parametrization (see~\Cref{sec:parametrization})}.
To enhance efficiency, this matrix inversion is performed via its Schur complement.
This complement requires computing $\massMatrix^{-1}$.
Similar to free dynamics, we achieve this through the Cholesky decomposition of $\massMatrix$, an approach typically employed in \glspl{imm}.

Similarly, derivatives of impulse/\rev{reset} dynamics \rev{$\resetFunc[p]$} involves inverting the same matrix but with:
\begin{gather}\label{eq:diff_imp_tau_p}
\begin{aligned}
\frac{\partial\eff}{\partial\params} &\coloneqq \left(\frac{\partial\text{ID}(\pos,\zeroVec,\vel^+ - \vel)}{\partial\dynParams} - \frac{\partial\text{ID}(\pos,\zeroVec,\zeroVec)}{\partial\dynParams}\right)\frac{\partial\dynParams}{\partial\params},\\
&= \left(\regMatrix(\pos,\zeroVec,\vel^+-\vel) - \regMatrix(\pos,\zeroVec,\zeroVec)\right)\frac{\partial\dynParams}{\partial\params},
\end{aligned}
\end{gather}
where $\vel^+\in\confTManif\subseteq\R^{\nq}$ is the post impact velocity.

\section{DDP with parametrized dynamics}\label{sec:PDDP}
Handling parametrized dynamics requires to introduce \rev{an} additional decision variable $\params$.
To exploit its \textit{temporal structure}, we begin by examining the optimality conditions through the lens of Bellman, ultimately leading to a parametrized Riccati recursion.
\rev{For readers not familar with numerical optimization, we provide a accompanying report in~\cite{smartrepot}.}

\subsection{Optimality conditions}
By examining the Bellman equation of~\Cref{eq:oe_complete}, specifically,
\begin{gather}\label{eq:bellman_eq}
\begin{aligned}
\valFunc\left(\state;\params\right\vert\ctrlMeas,\obsMeas) =& \min_{\state^\prime,\state,\ucertain;\params}\ell(\state,\ucertain;\params\vert\obsMeas)+\valFunc^\prime(\state^\prime;\params\vert\ctrlMeas,\obsMeas)\\
&\quad\text{s.t.}\quad \state^\prime = \dynFunc(\state,\ucertain;\params\vert\ctrlMeas),
\end{aligned}
\end{gather}
we break the optimal estimation problem into a sequence of subproblems.
The~\gls{kkt} point for each subproblem can be efficiently determined using the Newton method \rev{(i.e., $\nabla\mathbf{r}\,\delta\mathbf{r}=-\mathbf{r}$)}, yielding the following linear system of equations:
\begin{align}\label{eq:kkt}\nonumber
&\overbrace{\begin{bmatrix}
\lagHess[\state\state]& \lagHess[\state\ucertain] & \lagHess[\state\params] & \dynFunc[\state]^\transpose &\\
\lagHess[\state\ucertain]^\transpose & \lagHess[\ucertain\ucertain] & \lagHess[\ucertain\params] & \dynFunc[\ucertain]^\transpose &\\
\lagHess[\state\params]^\transpose &  \lagHess[\ucertain\params]^\transpose & \lagHess[\params\params] & \dynFunc[\params]^\transpose &   \valFunc[\state\params]^{\prime\transpose} \\
\dynFunc[\state] & \dynFunc[\ucertain] & \dynFunc[\params] & & - \mathbf{I} \\
& & \valFunc[\state\params]^{\prime} & - \mathbf{I} & \valFunc[\state\state]^{\prime} 
\end{bmatrix}}^{\rev{\nabla\mathbf{r}}}
\overbrace{\begin{bmatrix}
\delta \mathbf{x} \\ \delta \mathbf{w} \\ \delta \boldsymbol{\theta} \\ \mulpDynNext \\ \delta \mathbf{x}^{\prime} 
\end{bmatrix}}^{\rev{\delta\mathbf{r}}} =-
\overbrace{\begin{bmatrix}
\costGrad[\state] \\
\costGrad[\ucertain] \\
\costGrad[\params] + \valFunc[\params]^\prime \\
\new{\dynFeas} \\
 \valFunc[\state]^{\prime}
\end{bmatrix}}^{\rev{\mathbf{r}}}\\
&\text{with:}\\\nonumber
&\hspace{0.5em}\mulpDynNext \coloneqq \mulpDyn + \delta\mulpDyn, \hspace{5.8em}
\new{\dynFeas \coloneqq \dynFunc(\state, \ucertain; \params) \ominus \state^{\prime}},\\\nonumber
&\hspace{0.5em}\lagHess[\state\state] \coloneqq \costGrad[\state\state] + \valFunc[\state]^{\prime} \cdot \dynFunc[\state\state], \hspace{2em}
\lagHess[\state\ucertain] \coloneqq \costGrad[\state\ucertain] + \valFunc[\state]^{\prime} \cdot \dynFunc[\state\ucertain],\\\nonumber
&\hspace{0.5em}\lagHess[\state\params] \coloneqq \costGrad[\state\params] + \valFunc[\state]^{\prime} \cdot \dynFunc[\state\params], \hspace{2em}
\lagHess[\ucertain\ucertain] \coloneqq \costGrad[\ucertain\ucertain] + \valFunc[\state]^{\prime} \cdot \dynFunc[\ucertain\ucertain],\\\nonumber
&\hspace{0.5em}\lagHess[\ucertain\params] \coloneqq \costGrad[\ucertain\params] + \valFunc[\state]^{\prime} \cdot \dynFunc[\ucertain\params], \hspace{1.5em}
\lagHess[\params\params] \coloneqq \costGrad[\params\params] + \valFunc[\params\params]^{\prime} + \valFunc[\state]^{\prime} \cdot \dynFunc[\params\params],
\end{align}
\rev{where $\mathbf{r}$ represents the residual vector containing the gradient of the Lagrangian of~\Cref{eq:bellman_eq}. In~\Cref{eq:kkt}}, $\costGrad[\pvar]$, $\dynFunc[\pvar]$, are the first derivative of the cost and the system dynamics with respect to $\pvar$, with $\pvar$ a hypothetical decision variable that represents $\state$, $\ucertain$ or $\params$; $\costGrad[\pvar\pvar]$, $\dynFunc[\pvar\pvar]$ are the second derivatives; $\valFunc[\pvar]^{\prime}$, \rev{$\valFunc[\pvar\pvar]^{\prime}$ are} the gradient \rev{and Hessian} of the value function; $\dynFeas$ describe the infeasibility in the dynamics; \rev{$\mulpDyn$ is the Lagrange multiplier associated to the dynamics;} $\delta\state$, $\delta\ucertain$, $\delta\params$, $\delta\state^\prime$ and $\delta\mulpDyn$ provides the search direction computed for the primal and dual variables, respectively; and the \textit{prime} superscript is used to refer to the next node.

Similar to optimal control~\cite[Section 2.2]{mastalli22auro}, \rev{by inspecting the last row of~\Cref{eq:kkt} }we observe the presence of the Markovian structure, leading to $\new{\mulpDynNext = \valFunc[\state]^\prime + \begin{bmatrix} \valFunc[\state\state]^\prime & \valFunc[\state\params]^\prime\end{bmatrix}\begin{bmatrix}\delta\state^\prime \\ \delta\params\end{bmatrix}}$.
This connection becomes evident when augmenting the system's state with its parameters \rev{$\params$}, contributing to the condensation of the system of equations in~\Cref{eq:kkt} \rev{as}:
\begin{gather}\label{eq:condensed_kkt}
\begin{bmatrix}
\qualFunc[\state\state] & \qualFunc[\state\ucertain] & \qualFunc[\state\params] \\
\qualFunc[\state\ucertain]^\transpose & \qualFunc[\ucertain\ucertain] & \qualFunc[\ucertain\params] \\
\qualFunc[\state\params]^\transpose & \qualFunc[\ucertain\params]^\transpose & \qualFunc[\params\params]
\end{bmatrix}
\begin{bmatrix}
\delta\state\\
\delta\ucertain \\
\delta\params    
\end{bmatrix} = -
\begin{bmatrix}
\qualFunc[\state]\\
\qualFunc[\ucertain] \\
\qualFunc[\params]   
\end{bmatrix},
\end{gather}
where the $\qualFunc$'s terms represent the local approximation of the \textit{action-value function} whose expressions are:
\begin{align}\nonumber
&\small\qualFunc[\state\state] = \lagHess[\state\state] + \dynFunc[\state]^\transpose \valFunc[\state\state]^\prime \dynFunc[\state],
\hspace{2.1em}
\qualFunc[\ucertain\params] = \lagHess[\ucertain\params] + \dynFunc[\ucertain]^\transpose (\valFunc[\state\params]^\prime + \valFunc[\state\state]^\prime \dynFunc[\params]), \\\nonumber
&\small\qualFunc[\state\ucertain] = \lagHess[\state\ucertain] + \dynFunc[\state]^\transpose \valFunc[\state\state]^\prime \dynFunc[\ucertain], 
\hspace{1.7em}
\qualFunc[\state\params] = \lagHess[\state\params] + \dynFunc[\state]^\transpose (\valFunc[\state\params]^\prime + \valFunc[\state\state]^\prime \dynFunc[\params]), \\\nonumber
&\small\qualFunc[\ucertain\ucertain] = \lagHess[\ucertain\ucertain] + \dynFunc[\ucertain]^\transpose \valFunc[\state\state]^\prime \dynFunc[\ucertain], \hspace{1.2em}
\qualFunc[\params\params] = \lagHess[\params\params] + \dynFunc[\params]^\transpose (2\valFunc[\state\params]^\prime + \valFunc[\state\state]^\prime \dynFunc[\params]), \\\nonumber
&\small\qualFunc[\state] = \costGrad[\state] + \dynFunc[\state]^\transpose \new{\valFunc[\state]^+},
\hspace{4.9em}
\qualFunc[\params] = \costGrad[\params] + \valFunc[\params]^+ + \dynFunc[\params]^\transpose \new{\valFunc[\state]^+},\\
&\small\qualFunc[\ucertain] = \costGrad[\ucertain] + \dynFunc[\ucertain]^\transpose \new{\valFunc[\state]^+}
\end{align}
with $\new{\valFunc[\state]^+ \coloneqq \valFunc[\state]^\prime + \valFunc[\state\state]^\prime \dynFeas}$ and $\new{\valFunc[\params]^+ \coloneqq \valFunc[\params]^\prime + \valFunc[\state\params]^\prime \dynFeas}$ as the gradients of the value function after the deflection produced by the dynamics infeasibility $\dynFeas$ (see~\cite{mastalli-icra20}).
To exploit the \textit{parametric structure}, we then compute the estimation policy and value function as functions of the parameters and arrival state.

\subsection{Policy and value function}
\Cref{eq:condensed_kkt} \rev{computes the \gls{kkt} point of} a quadratic program that locally minimizes the Bellman equation~\Cref{eq:bellman_eq}, i.e.,
\begin{align}\label{eq:qp_bellman}
\delta \valFunc(\delta\state;\delta\params\vert\ctrlMeas,\obsMeas)& \simeq \\\nonumber
&\hspace{-4em}\min_{\delta\ucertain}
\frac{1}{2}
\begin{bmatrix}
1 \\
\delta\state \\
\delta\ucertain \\
\delta\params
\end{bmatrix}^\transpose
\begin{bmatrix}
& \qualFunc[\state]^\transpose & \qualFunc[\ucertain]^\transpose & \qualFunc[\params]^\transpose \\
\qualFunc[\state] & \qualFunc[\state\state] & \qualFunc[\state\ucertain] & \qualFunc[\state\params] \\
\qualFunc[\ucertain] & \qualFunc[\state\ucertain]^\transpose & \qualFunc[\ucertain\ucertain] & \qualFunc[\ucertain\params] \\
\qualFunc[\params] & \qualFunc[\state\params]^\transpose & \qualFunc[\ucertain\params]^\transpose & \qualFunc[\params\params]
\end{bmatrix}
\begin{bmatrix}
1 \\
\delta\state \\
\delta\ucertain \\
\delta\params
\end{bmatrix}.
\end{align}
\rev{Moreover, we compute $\delta\ucertain$ as a function of $\delta\state$ and $\delta\params$ from~\Cref{eq:condensed_kkt}}.
This leads to what we call the \textit{\gls{ddp} approach} for optimal estimation, \rev{as it computes an \textit{estimation policy} $\delta\ucertain$ given observations $(\ctrlMeas,\obsMeas)$, i.e.,}
\begin{gather}\label{eq:policy_w}
\delta\mathbf{w} = - \ffPolicy - \fbPolicy\delta\textbf{x} - \pPolicy\delta\params,
\end{gather} 
where $\ffPolicy=\qualFunc[\ucertain\ucertain]^{-1}\qualFunc[\ucertain]$, $\fbPolicy = \qualFunc[\ucertain\ucertain]^{-1}\qualFunc[\state\ucertain]^\transpose$ and $\pPolicy = \qualFunc[\ucertain\ucertain]^{-1}\qualFunc[\ucertain\params] $ are the feed-forward and feedback terms.
In this approach, when plugging the changes in the estimation into~\Cref{eq:qp_bellman}, we obtain a quadratic approximation of the \textit{optimal arrival cost} (a.k.a. value function) as follows
\begin{align}\nonumber
\delta\valFunc(\delta\state; \delta\params\vert\ctrlMeas,\obsMeas)& \simeq \Delta\valFunc[1] + \frac{\Delta\valFunc[2]}{2}\\
&\hspace{-4em} + \frac{1}{2}
\begin{bmatrix}
\delta\state\\
\delta\params
\end{bmatrix}^\transpose
\begin{bmatrix}
\valFunc[\state\state] & \valFunc[\state\params] \\
\valFunc[\state\params]^\transpose & \valFunc[\params\params]
\end{bmatrix}
\begin{bmatrix}
\delta\state\\
\delta\params
\end{bmatrix}
+ \begin{bmatrix}
\valFunc[\state]\\
\valFunc[\params]
\end{bmatrix}^\transpose
\begin{bmatrix}
\delta\state\\
\delta\params
\end{bmatrix},
\end{align}
where\vspace{-0.8em}
\begin{align}\label{eq:value_func_update}\nonumber
& \Delta\valFunc[1] = - \ffPolicy^\transpose \qualFunc[\ucertain], \quad\quad\quad\,\, \Delta\valFunc[2] = \ffPolicy^\transpose\qualFunc[\ucertain\ucertain]\ffPolicy, \\\nonumber
&\valFunc[\state] = \qualFunc[\state] - \qualFunc[\state\ucertain]\ffPolicy, \quad\quad\,
\valFunc[\params] = \qualFunc[\params] - \qualFunc[\ucertain\params]^\transpose\ffPolicy, \\\nonumber
&\valFunc[\state\state] = \qualFunc[\state\state] - \qualFunc[\state\ucertain] \fbPolicy, \quad  \valFunc[\params\params] = \qualFunc[\params\params] - \qualFunc[\ucertain\params]^\transpose\pPolicy, \\
&\valFunc[\state\params] = \qualFunc[\state\params] - \qualFunc[\state\ucertain]\pPolicy,
\end{align}
and $\qualFunc[\state\ucertain]\pPolicy=\fbPolicy^\transpose\qualFunc[\ucertain\params]$.

To compute the search directions $\delta\params$, we analyze the conditions of optimality for the initial node \rev{detailed below. 
Unlike the approach in~\cite{kobilarov2015optestddp}, 
our method calculates $\delta\params$ while simultaneously estimating the unknown arrival state $\delta\state[0]$} \REV{and accounting for dynamics infeasibilities}.

\subsection{Arrival state and parameters}
By substituting~\Cref{eq:policy_w} into the condensed \gls{kkt} equations of the arrival node in~\Cref{eq:condensed_kkt}, we obtain:
\begin{gather}\label{eq:initnode_kkt}
\new{\begin{bmatrix}
\valFunc[\state\state]^\bullet & \valFunc[\state\params]^\bullet \\
\valFunc[\state\params]^{\bullet\,\transpose} & \valFunc[\params\params]^\bullet
\end{bmatrix}
\begin{bmatrix}
\delta\state[0] \\ \delta\params
\end{bmatrix} = -
\begin{bmatrix}
\valFunc[\state]^\bullet \\ \valFunc[\params]^\bullet
\end{bmatrix}},
\end{gather}
where the \textit{bullet} superscript is used to refer to the arrival node (e.g., $\delta\state[0]\coloneqq\delta\state^\bullet$).
\rev{Next, we describe two methods for factorizing~\Cref{eq:initnode_kkt} namely the Schur-complement and nullspace methods.}

\subsubsection{Schur-complement \rev{method}}
\rev{It computes} both $\delta\params$ and $\delta\state[0]$ by factorizing~\Cref{eq:initnode_kkt} via the Schur-complement approach.
This results in the following expressions:
\begin{gather}\label{eq:policy_p_dx0}
\begin{array}{l}
\delta\params = -\ffPolicy[\params] -\fbPolicy[\params]\delta\state[0], \quad
\delta\state[0] \coloneqq -\valFunc[\state\initState]^{-1}\valFunc[\initState],
\end{array}
\end{gather}
where $\ffPolicy[\params]=\valFunc[\params\params]^{\bullet\, -1}\valFunc[\params]^{\bullet}$, $\fbPolicy[\params]=\valFunc[\params\params]^{\bullet\, -1}\valFunc[\state\params]^{\bullet\,\transpose}$ are the feed-forward and feedback terms of the parameters update \rev{$\delta\params$},
\begin{gather}\label{eq:value_func_arrival_node}
\begin{array}{l}
\valFunc[\initState] = \valFunc[\state]^\bullet - \valFunc[\state\params]^\bullet\ffPolicy[\params], \quad 
\valFunc[\state\initState] = \valFunc[\state\state]^\bullet - \valFunc[\state\params]^\bullet \fbPolicy[\params],
\end{array}
\end{gather}
are the derivatives of the value function associated to arrival node after updating the system's parameters, and $\valFunc[\state\initState]$ is of the Schur complement of~\Cref{eq:initnode_kkt}.
Now, we can rewrite the changes of the estimation policy as $\delta\ucertain = -\ffPolicyTotal - \fbPolicy\delta\state$,
where $\ffPolicyTotal=\ffPolicy+\pPolicy\delta\params$ represents the entire feed-forward term.

\subsubsection{Nullspace \rev{method}}
\rev{It parametrizes} the search direction for the parameters as $\delta\params = \mathbf{Y}\delta\params[\mathbf{y}] + \mathbf{Z}\delta\params[\mathbf{z}]$.
Here, $\mathbf{Z} \in \R^{\nparams \times n_z}$ is the nullspace basis of $\valFunc[\params\params]^{\bullet}$ and $\mathbf{Y} \in \R^{\nparams \times n_y}$ is its orthogonal matrix.
Then, by substituting this parametrization into the second term of~\Cref{eq:policy_p_dx0} and observing that $\valFunc[\params\params]^{\bullet} \mathbf{Z} = \zeroVec$, we obtain
\vspace{-3mm}
\begin{gather}\label{eq:policy_p_null}
\mathbf{Y}^\transpose \valFunc[\params\params]^\bullet \mathbf{Y}
\delta \params[\mathbf{y}] = 
-\overbrace{\mathbf{Y}^\transpose\valFunc[\params]^\bullet}^{\ffPolicy[\params \mathbf{y}]}
-\overbrace{\mathbf{Y}^\transpose\valFunc[\state\params]^{\bullet\,\transpose}}^{\fbPolicy[\params \mathbf{y}]}\delta\initState
,
\end{gather}
where we pre-multiply by $\mathbf{Y}^\transpose$ to ensure the squareness of $\valFunc[\params\params]\mathbf{Y}$.
\Cref{eq:policy_p_null} is then solved efficiently using a Cholesky decomposition \rev{as $\mathbf{Y}^\transpose \valFunc[\params\params]^\bullet \mathbf{Y}$ is a positive definite symmetric matrix}.
Finally, we recover the search direction for the parameters as $\delta\params = \mathbf{Y}\delta\params[\mathbf{y}]$ \rev{and  $\delta\initState$ from the second term of~\Cref{eq:policy_p_dx0}}.

\REV{The asymptotic computational complexity of our Riccati recursion is $\mathcal{O}(\mathcal{N}(\nx+\nx)^{3}+\nparams^{3})$ flops,\footnote{\rev{For more details on algorithmic complexity of Riccati recursion, we refer to~\cite{GianlucaFrison2013}.}} which \finalrev{shows potential for} real-time applications.
This is because full-dynamics \gls{mpc} algorithms with similar complexity have demonstrated the capability to run at high frequencies (e.g.,~\cite{mastalli2022agile}).
}

\subsection{Forward rollouts \rev{and merit function}}
The search directions obtained from \Cref{eq:policy_w,eq:policy_p_dx0,eq:policy_p_null} provide updates for the system's uncertainties, parameters, and arrival state.
To determine the step length along these directions, we perfom a forward rollout of the disturbed system's dynamics and evaluate the associated costs.
\rev{Below, we describe our novel approach inspired by optimal control literature, known as multiple shooting.}
Let's start by describing how we update the arrival state and parameters first.

\subsubsection{Arrival state and parameters update}
We follow a line-search approach \rev{to update} the arrival state and system's parameters, i.e.,
\vspace*{-0.5\baselineskip}
\begin{gather}\label{eq:try_x0_p}
\stateNext[0] = \state[0]\oplus\stepLength\delta \state[0],\quad
\paramsNext = \params+\stepLength\delta\params,
\end{gather}
where $\stepLength\in(0,1]$ denotes the step length and the \textit{plus} superscript is used to denote the next guess under evaluation.


\subsubsection{Multiple shooting rollout}
\rev{Building upon classical line-search procedures in optimal control (see~\cite[Chapter~3]{nocedal-optbook})}, \rev{our multiple shooting approach} tries a linear rollout and updates the gaps based on nonlinear shoots, i.e.,
\begin{gather}
\begin{aligned}
\label{eq:multiple_shooting}
\ucertainNext[k] &= \ucertain[k] + \stepLength \delta\ucertain[k],\\
\stateNext[k+1] &= \state[k+1]\oplus\stepLength\delta\state[k+1],\\
\dynFeasNext[k+1] &= \dynFunc(\stateNext[k], \ucertainNext[k]; \paramsNext[k]) \ominus \stateNext[k+1],
\end{aligned}
\end{gather}
for \REV{$k\in\{0,\ldots,N\} \subset\N$}.
\rev{Compared to the feasibility-driven rollout in \cite{mastalli-icra20}, multiple shooting rollouts can be broken into two parallel for loops: one for the update of the uncertainties $\ucertainNext[k]$ and state $\stateNext[k+1]$, and another one for the update of the dynamics infeasibilities $\dynFeasNext[k+1]$ and costs $\ell_{k}(\stateNext[k], \ucertainNext[k])$.}
Below, we describe how we evaluate the goodness of a given step length $\stepLength$.

\subsubsection{Merit function}
\rev{We develop a merit function inspired by~\cite{mastalli-invdynmpc}.
Specifically, it computes the expected cost reduction as a local approximation of the value function while being aware of the dynamics infeasibilities.
More details can be found in our technical report~\cite{smartrepot}.}
Finally, we utilize (1) a nonmonotone step acceptance strategy based on Armijo conditions and (2) a Levenberg-Marquardt scheme \rev{as in}~\cite{mastalli-invdynmpc}.

\section{Results}\label{sec:reuslts}
We evaluate our multi-contact inertial \REV{parameters} estimation and localization framework \rev{as follows}. 
First, we compare the numerical performance of our exponential eigenvalue parametrization \REV{in~\Cref{eq:exp_eig}} against the log-Cholesky \REV{in~\Cref{eq:logchol_param}} approach. 
Second, we analyze the effect of \rev{multiple} shooting approaches in optimal estimation. 
Third, we experimentally validate the advantages of updating inertial parameters for carrying unknown payloads.
Finally, we show the importance of using hybrid dynamics for localization in \rev{agile maneuvers}.
\rev{We refer the reader to the video, where the benefits of our nullspace resolution are showcased.}

\subsection{Log-Cholesky vs Exponential Eigenvalue}
\rev{This} numerical analysis encompasses four diverse robotics systems: the Kinova arm, \rev{the Hector} quadrotor, the ANYmal quadruped, and the Talos biped.
In each scenario, we generated dynamically-consistent trajectories 
\rev{by integrating observations from simulated joint encoders $(\posMeas[j],\velMeas[j])$, torso's IMU data (linear acceleration $\accMeas[i]$, gyroscopic velocities $\angVelMeas[i]$, and orientations), and joint \REV{efforts} $\ctrlMeas$.}
To ensure a fair comparison between parameterizations, we omitted the parameter regularization term $\|\params - \paramsMean\|^2_{\covariance_{\params}^{-1}}$ \REV{from \Cref{eq:oe_problem}}.
Moreover, we initialized the inertial parameters for all the bodies with the same error of $70\%$.
%
\REV{For the ANYmal and Talos robots, we included the contact constraints in~\Cref{eq:contact_constraint}.
All the cases incorporated the multibody dynamics described in~\Cref{eq:dynamics}.}

\begin{figure}[b]\centering
\includegraphics[width = 1.\linewidth]{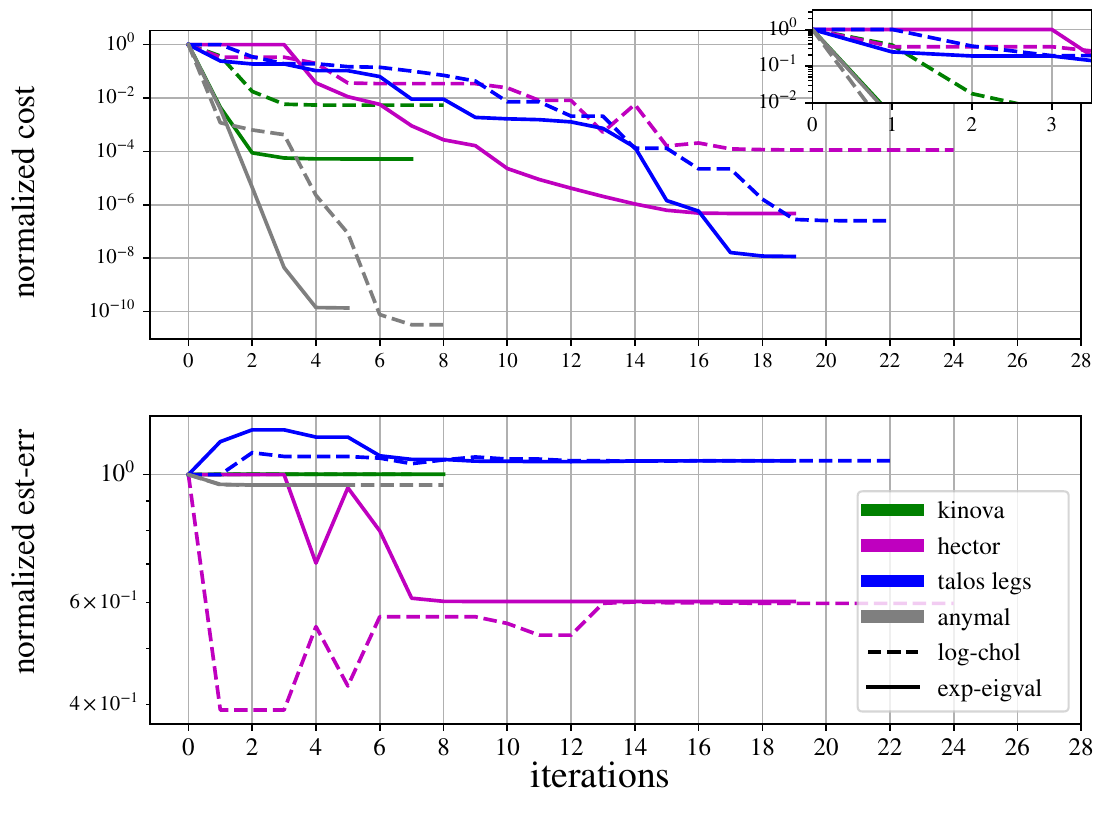}
\caption{Top: Local convergence for both parametrizations, showing better convergence the \textit{exponential eigenvalue} (exp-eigval) parametrization in all cases.
Bottom: Evolution of the estimation error, computed as the $\ell_1$-norm of difference between the optimized trajectory and the nominal one.
Both parametrizations converge to the same local minima.}
\label{fig:parametrizations}
\end{figure}

In each scenario illustrated in~\Cref{fig:parametrizations}, both parametrization converge to the same estimated trajectory, resulting in \rev{identical} estimation errors \rev{(\Cref{fig:parametrizations}-bottom)}.
Nevertheless, the exponential eigenvalue parametrization demonstrated superior performance, characterized by faster convergence rates and often lower cost values \rev{(\Cref{fig:parametrizations}-top)}.
This can be attributed to its lower degree of nonlinearity compared to the log-Cholesky parametrization. 
Furthermore, the exponential eigenvalue parametrization offers a direct physical interpretation, \rev{which is particularly advantageous} when partial knowledge of the rotational inertia is available. 
This \rev{is because it enables us to easily tune} covariances in known inertial parameters \rev{$\covariance_{\params}$}, \rev{enhacing results} in \rev{our} trials with the Go1 robot.

\subsection{Numerical effect of rollout strategies}
To evaluate the numerical impact of our feasibility-driven \REV{~\citeeq{smartrepot}{Eq.~(30)}} and multiple shooting rollouts \REV{in~\Cref{eq:multiple_shooting}}, we randomly generated $100$ initial \rev{guesses} and compared their convergence, cost and estimation errors against single shooting approaches \rev{(i.e., approach developed in~\cite{kobilarov2015optestddp})}.
In~\Cref{tab:rollouts}, we observe that multiple shooting rollouts reduced the costs or estimation errors across all cases.
Moreover, single shooting rollouts have a reduced basin of attraction to local minima and can encounter difficulties in convergence for challenging cases.
\rev{Multiple shooting rollouts enabled us to handle the agile manuevers in~\Cref{fig:snapshots}.}

\begin{table}[t]\centering
\caption{Rollout's local convergence computed from $100$ randomly initial guesses on three robotics systems. \rev{\textit{P-value} obtained from an ANOVA analysis \REV{after applying Bonferroni correction}.}}
\label{tab:rollouts}
\ra{1.1}
\begin{tabular}{@{} l @{\hspace{1.\tabcolsep}} l @{\hspace{1.\tabcolsep}} ccc}
&  & Iterations & Cost $[\cdot 10^{-1}]$ & Error [$\ell_\infty$-norm]\\
\hline
\multirow{3}{*}{\textit{Kinova}} 
& single &   $39.7 \pm 11.2$ & $2.78 \pm 0.89$ & $3.11 \pm 0.56$\\
& feasible & $\mathbf{44.7 \pm 11.6}$ & $1.16 \pm 0.42$ & $3.12 \pm 0.01$\\
& multiple & $99.4 \pm 71.5$ & $\mathbf{0.61 \pm 0.42}$ & $\mathbf{2.67 \pm 0.06}$\\
&\rev{\it{p-value}} & \REV{$1.19\cdot10^{-60}$} & \REV{$1.64\cdot10^{-17}$} & \REV{$1.08\cdot10^{-5}$}\\
\midrule
\multirow{3}{*}{\textit{Quadrotor}} 
& single &   $140.9 \pm 65.1$ & $0.12 \pm 0.04$ & $10.74 \pm 0.0023$\\
& feasible & $150.1 \pm 59.3$ & $0.26 \pm 0.11$ & $10.74 \pm 0.0013$\\
& multiple &  $\mathbf{16.9 \pm 19.1}$ & $\mathbf{0.07 \pm 1.31}$ & $\mathbf{10.74 \pm 0.0002}$\\
&\rev{\it{p-value}} & \REV{$3.51\cdot10^{-5}$} & \REV{$2.04\cdot10^{-85}$} & \REV{$1.26\cdot10^{-22}$}\\
\midrule
\multirow{3}{*}{\textit{ANYmal}} 
& single &   \tikzxmark & \tikzxmark & \tikzxmark \\
& feasible & $198.3 \pm 16.6$ & $0.22 \pm 0.08$ & $6.12 \pm 0.92$\\
& multiple & $\mathbf{14.1 \pm 14.8}$ & $\mathbf{0.01 \pm 1.65}$ & $\mathbf{6.02 \pm 0.22}$\\
&\rev{\it{p-value}} & \REV{$2.07\cdot10^{-79}$} & \REV{$5.64\cdot10^{-4}$} & \REV{$7.41\cdot10^{-4}$}\\
\midrule
\end{tabular}
\tikzxmark algorithm does not find a solution within 200 iterations.\vspace{-1em}
\end{table}
%
\subsection{Validation on the Go1 robot}\label{sec:go1_exps}
\begin{figure}[b]
  \href{\video&t=77}{\includegraphics[width = 0.98\linewidth]{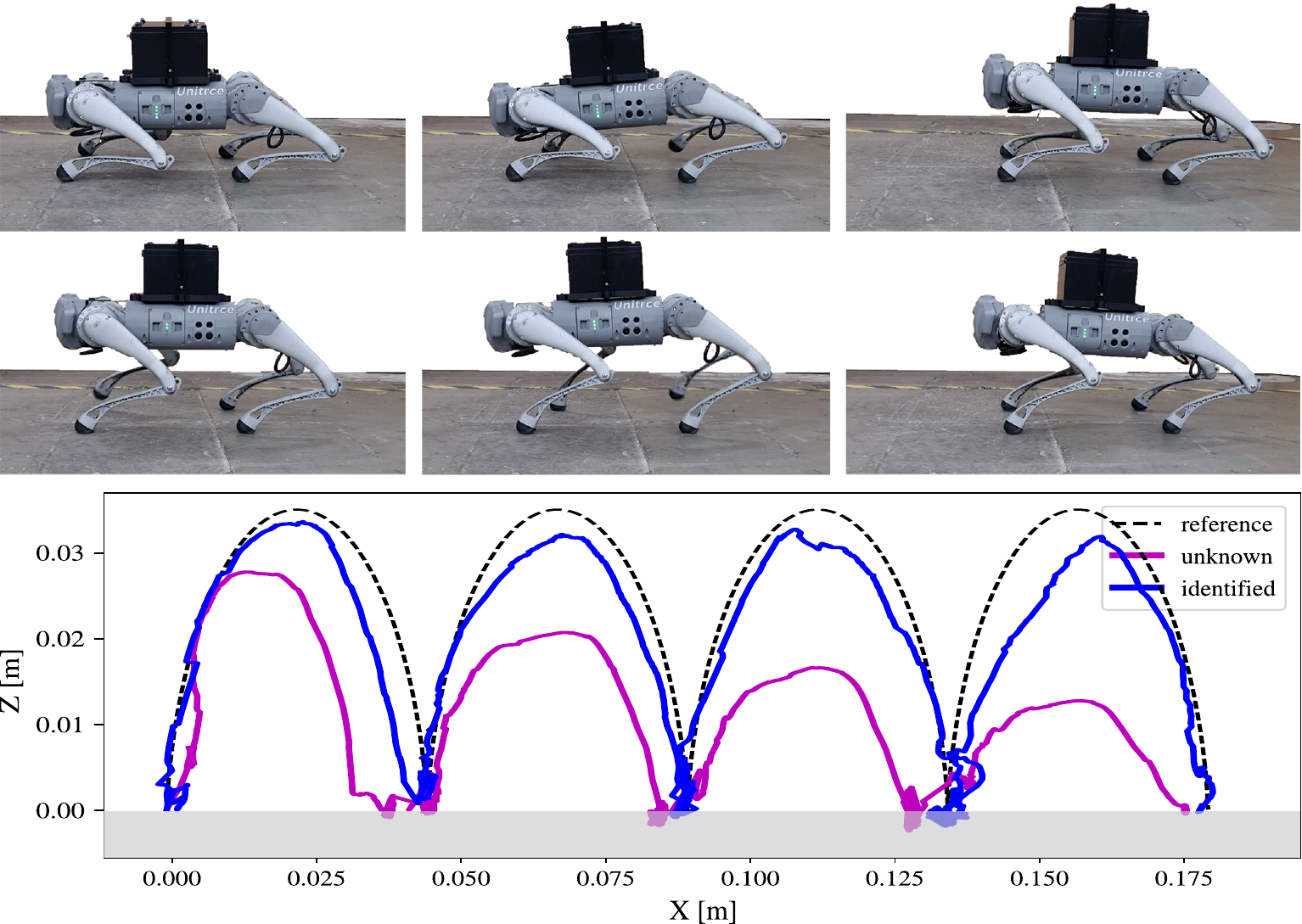}}
  \caption{\REV{Top: Go1 performing four walking gaits while carrying an unknown payload.
  In the top snapshots, Go1 struggled to maintain its posture due to model mismatch in the~\gls{mpc}.
  In the bottom snapshots, after correctly estimating the robot's payload, Go1 confidently maintained its nominal posture. Bottom: Foot-swing tracking, the foot position is obtained from the VICON system, ignoring any localization effects.
  Our~\gls{mpc} encountered difficulties in tracking the reference swing-foot trajectory (dashed line) when the payload was unknown (\rev{magenta} line).
  Specifically, the Go1 robot failed to reach the desired step height, an error accumulated over time.
In contrast, our~\gls{mpc} improved its foot-swing tracking performance when our estimator identified the payload (\rev{blue} line).}}
  \label{fig:go1_experiment}
\end{figure}

We validated our multi-contact inertial estimation and localization in \rev{experimental} trials with the Go1 robot.
To showcase its capability in inertial \rev{parameters} estimation, we added an unknown payload of \SI{7.2}{\kilogram} to the robot's torso.
The observations \rev{were} recorded when the Go1's predictive controller~\rev{\cite{mastalli2022agile}} was unaware of the additional payload (\Cref{fig:go1_experiment}-top).

We integrated a variety of \rev{proprioceptive measurements} \REV{$\obsMeas\in\obsManif$ in ~\Cref{eq:oe_problem}} with a step integration interval of \SI{1}{\milli\second}.
First, we included joint positions $\rev{\posMeas[j]}\in\R^{12}$ \rev{measured from encoders} and numerically differentiated them to account for joint velocities $\rev{\velMeas[j]}\in\R^{12}$.
Second, we derived Go1's torso rotation from the IMU's sensor fusion algorithm.
This was incorporated with the following observation model $\|\rotMatrixMeas[b]\ominus\rotMatrix_b\|^2_{\rotMatrixCov^{-1}}$.
Here, $\rotMatrixMeas[b]\ominus\rotMatrix_b$ denotes the \SO[3] inverse composition between the measured body orientation $\rotMatrixMeas[b]$ and the estimated one $\rotMatrix_b$ and $\covariance_\rotMatrix\in\Sgroup[3]$ represents the covariance estimated by the sensor fusion algorithm.
Third, we pre-integrated the IMU's linear acceleration to obtain the torso's linear velocities.
This observation also encompasses the IMU's angular velocities, both of them are expressed in local coordinates for efficiency reasons.
\rev{We estimated the contact sequence (see \Cref{fig:estimationGO1_pos}-top) using \textsc{Pronto}'s default method, which is based on the contact forces estimated through a standard algorithm that utilizes the dynamic model in~\Cref{eq:dynamics,eq:contact_constraint}.}
Lastly, we incorporated the torque commands $\ctrlMeas\in\R^{12}$ applied by the \rev{\gls{mpc}} controller.

\subsubsection{Estimating an unknown payload} \rev{To showcase the importance of a correct estimation of the inertial parameters we ran our optimal estimator offline.}
Within the estimation horizon of \SI{14}{\second}, it accurately determined the total payload, equivalent to a mass of \SI{7.364}{\kilogram} \rev{after $27$ iterations.}
Subsequently, we updated the inertial parameters of Go1's torso in its full-dynamics~\gls{mpc} \rev{controller}.
The impact of this update is vividly illustrated in~\Cref{fig:go1_experiment}-top.
Concretely, it enabled the Go1 robot to maintain a stable posture during  walking, emphasizing the significance of correcting model mismatches even for \rev{our} fast~\gls{mpc} running at \SI{50}{\hertz}.

\begin{table}[t]\centering
\caption{\REV{Identification errors of the unknown payload in the Go1 experimental trials.
The window size for the Savitzky–Golay filter used in the least squares (LS) method is indicated in braces.}}
\label{tab:sysID-leastsqr}
\begin{tabular}{@{} c @{\hspace{1.\tabcolsep}} ccc}
\REV{Gaussian noise $\sigma$} & \REV{LS [100ms]} & \REV{LS [10ms]} & \REV{OE (ours)}\\
\hline
\REV{$0.0$}    &   \REV{$102.20\%$} & \REV{$\mathbf{3.87\%}$} & \REV{$\mathbf{2.27\%}$}\\
\REV{$0.005$}  &   \REV{$101.51\%$} & \REV{$\mathbf{8.9\%}$} & \REV{$\mathbf{2.28\%}$}\\
\REV{$0.01$}   &   \REV{$102.06\%$} & \REV{$38.48\%$       } & \REV{$\mathbf{2.54\%}$}\\
\REV{$0.1$}    &   \REV{$34.98\%$ } & \REV{$170.48\%$      } & \REV{$\mathbf{4.67\%}$}\\
\midrule
\end{tabular}
\vspace*{-2.5\baselineskip}
\end{table}
The payload estimation significantly influences the tracking capabilities of the swing foot, as depicted in~\Cref{fig:go1_experiment}-bottom.
A comparison between swing trajectories with and without \rev{estimated} inertial \rev{parameters} reveals the crucial role it plays.
In the absence of accurate inertial parameters, the~\gls{mpc} struggled to track reference swing trajectories, resulting in a progressively decaying motion.
Instead, when the inertia was estimated, the~\gls{mpc} adeptly tracked the swing trajectory.
\REV{
\subsubsection{Comparison of our optimal estimator with least squares}
To evaluate the performance of our inertial estimation method, we compared its accuracy with the regression (least squares) approach proposed in~\cite{rucker2022smooth}.
\Cref{tab:sysID-leastsqr} presents the identification errors for both methods.
For the least squares method, accelerations for the joint-torque regressor were computed via numerical differentiation and processed using a filtering technique.
Results are provided for two filter configurations: a suggested \SI{100}{\milli\second} window in~\cite{rucker2022smooth} and an improved \SI{10}{\milli\second} window.
Each row in the table includes results for datasets with a possible Gaussian noise added to all observations.}

\REV{The least squares method exhibits a high sensitivity to acceleration values, making the choice of filter a critical factor in the accuracy of inertial estimation.
This sensitivity underscores the advantages of our Bayesian approach, which also handles localization uncertainties.}

\subsubsection{Localization with hybrid dynamics}\label{sec:go1_localization_results}
\begin{figure}
\centering
\href{\video&t=106}{\includegraphics[width = 1.\linewidth]{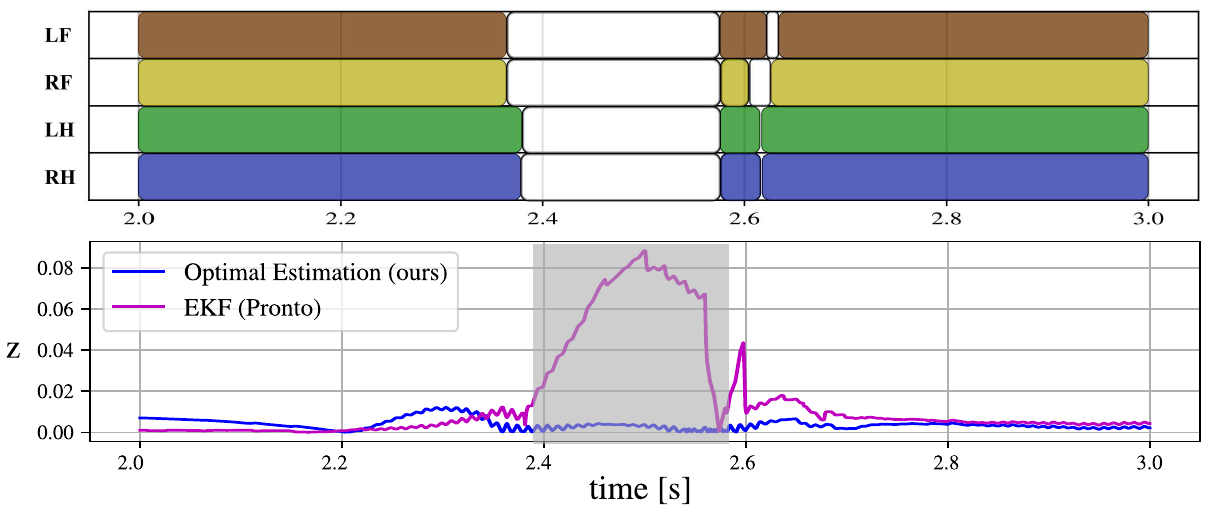}}
\caption{\rev{Contact estimation of contact and} Go1's localization errors when performing a jump.
Our optimal estimator (blue lines) exhibits a smaller estimation error compared to \textsc{Pronto}.}
\label{fig:estimationGO1_pos}
\end{figure}

\rev{To show the importance of a dynamic model in estimation, w}e compared our multi-contact localization approach against \textsc{Pronto}~\cite{fallon14pronto}--a widely adopted localization framework for legged robots based on~\gls{ekf} and kinematics.
\textsc{Pronto} and \REV{our} optimal estimator were configured to integrate \rev{the same} proprioceptive information (namely IMU and encoders), contact estimation, integration step, and covariances.
To provide a holistic evaluation, we employed \textsc{Pronto} to close the loop with our~\gls{mpc}.
\Cref{fig:estimationGO1_pos}-bottom shows localization errors for both estimators, measured \REV{with a} motion capture \REV{system}.

The inclusion of hybrid dynamics significantly improved localization accuracy, especially in the presence of a dynamic maneuver like a jump, \rev{surpassing the estimation accuracy of} \textsc{Pronto}'s estimation.
This achievement is \rev{possible} because introducing dynamics \rev{effectively} corrects \rev{nearly} all IMU drifts\rev{, highlighting the critical role of relying on the robot’s dynamics to mitigate and correct this drift}.
\rev{We also anticipate similar performance compared to traditional factor graph methods when relying on kinematic models.
This is because kinematic models struggle to accurately predict flying motions.
}

\section{Conclusion}\label{sec:conclusion}
We introduced (i) a multiple shooting optimal estimation algorithm tailored for multi-contact inertial estimation and localization and (ii) a novel smooth manifold with local submersion, named exponential eigenvalue, to ensure the full physical consistency of inertial parameters.
Our exponential eigenvalue manifold was compared against the singularity-free log-Cholesky manifold, demonstrating improved convergence attributed to its reduced degree of nonlinearity.
However, it presented singularities with symmetrical inertias.
To address this, we proposed a nullspace approach, which handles these rank deficiencies.
Additionally, our multiple shooting rollout demonstrated superior numerical behavior compared to existing methods, resulting in better estimations.
We demonstrated the effectiveness of our framework in complex scenarios such as humanoid brachiation and backflips, \REV{achieving higher accuracy than a conventional least squares approach}. 
Practical benefits were further illustrated through experimental trials on the Go1 robot.
Future work will involve an open-source C++ implementation.

\bibliographystyle{IEEEtran}
\bibliography{references}

\end{document}

%% file: math_commands.tex
\usepackage{amssymb}

\newcommand{\rev}[1]{\textcolor{black}{#1}}
\newcommand{\new}[1]{\textcolor{black}{#1}}
\newcommand{\REV}[1]{\textcolor{black}{#1}}

\newcommand{\finalrev}[1]{\textcolor{black}{#1}}

\newcommand{\citeeq}[2]{\cite[#2]{#1}}
\newcommand{\dynFunc}[1][]{\mathbf{f}_{#1}}
\newcommand{\obsFunc}[1][]{\mathbf{h}_{#1}}

\newcommand{\resetFunc}[1][]{\boldsymbol{\Delta}_{#1}}

\newcommand{\nx}{{n_{x}}} 
\newcommand{\nparams}{{n_{\theta}}} 
\newcommand{\nq}{{n_{q}}} 
\newcommand{\ntau}{{n_{u}}} 
\newcommand{\nz}{{n_{z}}} 
\newcommand{\nc}{{n_{c}}} 
\newcommand{\nb}{{n_{b}}} 

\newcommand{\state}[1][]{\mathbf{x}_{#1}} 
\newcommand{\ctrl}[1][]{\mathbf{u}_{#1}} 
\newcommand{\ucertain}[1][]{\mathbf{w}_{#1}} 
\newcommand{\params}[1][]{\boldsymbol{\theta}_{#1}} 
\newcommand{\stateSeq}{\state[s]}
\newcommand{\ucertainSeq}{\ucertain[s]}

\newcommand{\obsMeas}[1][]{\mathbf{\hat{z}}_{#1}} 
\newcommand{\posMeas}[1][]{\mathbf{\hat{q}}_{#1}} 
\newcommand{\velMeas}[1][]{\mathbf{\hat{v}}_{#1}} 
\newcommand{\accMeas}[1][]{\mathbf{\hat{q}}_{#1}} 
\newcommand{\angVelMeas}[1][]{\boldsymbol{\hat{\omega}}_{#1}} 
\newcommand{\forcelMeas}[1][]{\boldsymbol{\hat{\lambda}}_{#1}} 
\newcommand{\ctrlMeas}[1][]{\mathbf{\hat{u}}_{#1}} 
\newcommand{\covariance}{\boldsymbol{\Sigma}}
\newcommand{\stateMean}[1][]{\mathbf{\bar{x}}_{#1}} 
\newcommand{\stateCov}[1][]{\covariance_{\mathbf{x}_{#1}}} 
\newcommand{\ucertainCov}[1][]{\covariance_{\mathbf{w}_{#1}}} 
\newcommand{\paramsMean}[1][]{\boldsymbol{\bar{\theta}}_{#1}} 
\newcommand{\paramsCov}[1][]{\covariance_{\boldsymbol{\theta}_{#1}}} 
\newcommand{\obsCov}[1][]{\covariance_{\mathbf{\hat{z}}_{#1}}} 

\newcommand{\pos}[1][]{\mathbf{q}_{#1}} 
\newcommand{\vel}[1][]{\mathbf{v}_{#1}} 
\newcommand{\acc}[1][]{\mathbf{\dot{v}}_{#1}} 
\newcommand{\eff}[1][]{\boldsymbol{\tau}_{#1}} 
\newcommand{\wrch}[1][]{\boldsymbol{\lambda}_c}
\newcommand{\rotMatrixMeas}[1][]{\mathbf{\hat{\rotMatrix}}_{#1}}
\newcommand{\rotMatrixCov}{\covariance_\rotMatrix}

\newcommand{\N}{\mathbb{N}}
\newcommand{\R}{\mathbb{R}} 
\newcommand{\confManif}{\mathcal{Q}} 
\newcommand{\confTManif}[1][\pos]{\mathcal{T}_{#1}\confManif} 
\newcommand{\stateManif}{\mathcal{X}} 
\newcommand{\stateTManif}[1][\state]{\mathcal{T}_{#1}\stateManif} 
\newcommand{\obsManif}{\mathcal{Z}} 
\newcommand{\inertialManif}{\mathcal{I}}
\newcommand{\rotMatrix}{\mathbf{R}}
\newcommand{\rotMatrixParam}{\boldsymbol{\omega}}
\newcommand{\pseudoInertiaManif}[1][]{\mathcal{S}_{++}^{#1}}

\newcommand{\dynParams}[1][]{\boldsymbol{\pi}_{#1}} 
\newcommand{\massMatrix}{\mathbf{M}}
\newcommand{\coriolisGravTerm}{\mathbf{h}}
\newcommand{\contactJac}{\mathbf{J}_{c}}
\newcommand{\contactForce}{\boldsymbol{\lambda}_{c}}
\newcommand{\contactAcc}{\mathbf{a}_{c}}
\newcommand{\massDensity}{\rho}
\newcommand{\mass}{m}
\newcommand{\lever}{\mathbf{h}}
\newcommand{\barycenter}{\mathbf{c}}
\newcommand{\rotInertia}{\mathbf{I}}
\newcommand{\rotInertiaBarycenter}{\rotInertia_c}
\newcommand{\eigInertia}{\mathbf{D}}
\newcommand{\regMatrix}{\mathbf{Y}}
\newcommand{\pseudoInertia}[1][]{\mathbf{J}_{#1}}
\newcommand{\covInertia}[1][]{\boldsymbol{\Sigma}_{#1}}
\newcommand{\paramsMap}[1][]{\boldsymbol{\psi}_{#1}}
\newcommand{\secMomentMass}{\mathbf{L}}

\newcommand{\initState}{\state[0]} 
\newcommand{\pvar}[1][]{\mathbf{p}_{#1}}

\newcommand{\valFunc}[1][]{\mathcal{V}_{#1}}
\newcommand{\qualFunc}[1][]{\mathbf{Q}_{#1}}
\newcommand{\costGrad}[1][]{\boldsymbol{\ell}_{#1}}
\newcommand{\lagHess}[1][]{\mathcal{L}_{#1}}
\newcommand{\mulpDyn}[1][]{\boldsymbol{\xi}_{#1}}
\newcommand{\mulpDynNext}[1][]{\boldsymbol{\xi}_{#1}^{\mathbf{+}}}
\newcommand{\ffPolicy}[1][]{\mathbf{k}_{#1}}
\newcommand{\ffPolicyTotal}[1][]{\mathbf{\bar{k}}_{#1}}
\newcommand{\fbPolicy}[1][]{\mathbf{K}_{#1}}
\newcommand{\pPolicy}[1][]{\mathbf{P}_{#1}}
\newcommand{\stepLength}{\alpha}
\newcommand{\dynFeas}[1][]{\mathbf{\bar{f}}_{#1}}
\newcommand{\stateNext}[1][]{\mathbf{x}_{#1}^{\mathbf{+}}}
\newcommand{\ucertainNext}[1][]{\mathbf{w}_{#1}^{\mathbf{+}}}
\newcommand{\paramsNext}[1][]{\boldsymbol{\theta}_{#1}^{\mathbf{+}}}
\newcommand{\dynFeasNext}[1][]{\mathbf{\bar{f}}_{#1}^\mathbf{+}}

\newcommand{\dParam}[1][]{{d_{#1}}}
\newcommand{\sParam}[1][]{{s_{#1}}}
\newcommand{\tParam}[1][]{{t_{#1}}}

\newcommand{\transpose}{\intercal}
\newcommand{\zeroVec}[1][]{\mathbf{0}_{#1}}
\newcommand{\eyeMatrix}[1][]{\mathbf{1}_{#1}}
\newcommand{\zeroMatrix}[1][]{\mathbf{0}_{#1}}
\newcommand{\upperTriangMatrix}[1][]{\mathbf{U}_{#1}}
\newcommand{\skewMatrix}{\mathbf{S}}
\newcommand{\so}[1][]{\mathfrak{s0}(#1)}
\newcommand{\SO}[1][]{SO(#1)}
\newcommand{\Sgroup}[1][]{S(#1)}
\newcommand{\Tr}[1][]{\text{Tr}(#1)}

%% file: main.bbl
\begin{thebibliography}{10}
\providecommand{\url}[1]{#1}
\csname url@samestyle\endcsname
\providecommand{\newblock}{\relax}
\providecommand{\bibinfo}[2]{#2}
\providecommand{\BIBentrySTDinterwordspacing}{\spaceskip=0pt\relax}
\providecommand{\BIBentryALTinterwordstretchfactor}{4}
\providecommand{\BIBentryALTinterwordspacing}{\spaceskip=\fontdimen2\font plus
\BIBentryALTinterwordstretchfactor\fontdimen3\font minus \fontdimen4\font\relax}
\providecommand{\BIBforeignlanguage}[2]{{%
\expandafter\ifx\csname l@#1\endcsname\relax
\typeout{** WARNING: IEEEtran.bst: No hyphenation pattern has been}%
\typeout{** loaded for the language `#1'. Using the pattern for}%
\typeout{** the default language instead.}%
\else
\language=\csname l@#1\endcsname
\fi
#2}}
\providecommand{\BIBdecl}{\relax}
\BIBdecl

\bibitem{optest2004textbook}
J.~Crassidis and J.~Junkins, \emph{Optimal Estimation of Dynamic Systems}.\hskip 1em plus 0.5em minus 0.4em\relax Chapman and Hall, 2004.

\bibitem{foster-tro17}
C.~Forster, L.~Carlone, F.~Dellaert, and D.~Scaramuzza, ``\href{https://ieeexplore.ieee.org/document/7557075}{On-Manifold Preintegration for Real-Time Visual--Inertial Odometry},'' \emph{IEEE Trans. Rob. (T-RO)}, vol.~33, 2017.

\bibitem{wish-ral23}
D.~Wisth, M.~Camurri, and M.~Fallon, ``\href{https://ieeexplore.ieee.org/document/9852710}{VILENS: Visual, Inertial, Lidar, and Leg Odometry for All-Terrain Legged Robots},'' \emph{IEEE Rob. Autom. Lett. (RA-L)}, vol.~39, 2023.

\bibitem{wieber-fmbr05}
P.-B. Wieber, ``\href{https://link.springer.com/chapter/10.1007/978-3-540-36119-0_20}{Holonomy and nonholonomy in the dynamics of articulated motion},'' in \emph{{Fast Motions in Biomechanics and Robotics}}, 2005.

\bibitem{orsolino-ral18}
R.~{Orsolino}, M.~{Focchi}, C.~{Mastalli}, H.~{Dai}, D.~{Caldwell}, and C.~{Semini}, ``\href{https://doi.org/10.1109/LRA.2018.2836441}{Application of Wrench-Based Feasibility Analysis to the Online Trajectory Optimization of Legged Robots},'' \emph{IEEE Rob. Autom. Lett. (RA-L)}, vol.~3, 2018.

\bibitem{gabay82jota}
D.~Gabay, ``\href{https://hal.inria.fr/inria-00076552/file/RR-0009.pdf}{Minimizing a differentiable function over a differential manifold},'' \emph{J. Optim. Theory Appl.}, vol.~37, 1982.

\bibitem{mastalli-icra20}
C.~Mastalli, R.~Budhiraja, W.~Merkt, G.~Saurel, B.~Hammoud, M.~Naveau, J.~Carpentier, L.~Righetti, S.~Vijayakumar, and N.~Mansard, ``\href{https://cmastalli.github.io/publications/crocoddyl20unpub.html}{Crocoddyl: An Efficient and Versatile Framework for Multi-Contact Optimal Control},'' in \emph{IEEE Int. Conf. Rob. Autom. (ICRA)}, 2020.

\bibitem{sola2018micro}
J.~Sola, J.~Deray, and D.~Atchuthan, ``\href{https://arxiv.org/abs/1812.01537}{A micro Lie theory for state estimation in robotics},'' \emph{arXiv preprint arXiv:1812.01537}, 2018.

\bibitem{betts-bookoptctrl}
J.~T. Betts, \emph{\href{https://dl.acm.org/doi/book/10.5555/1734063}{Practical Methods for Optimal Control and Estimation Using Nonlinear Programming}}.\hskip 1em plus 0.5em minus 0.4em\relax Cambridge University Press, 2009.

\bibitem{gill-siam05}
P.~E. Gill, W.~Murray, and M.~A. Saunders, ``\href{https://dl.acm.org/doi/10.1137/S0036144504446096}{SNOPT: An SQP Algorithm for Large-Scale Constrained Optimization},'' \emph{SIAM Rev.}, vol.~12, 2005.

\bibitem{byrd-knitro06}
R.~H. Byrd, J.~Nocedal, and R.~A. Waltz, ``\href{http://citeseerx.ist.psu.edu/viewdoc/summary?doi=10.1.1.126.425}{KNITRO: An integrated package for nonlinear optimization},'' in \emph{Lrg. Scal. Nonlin. Opt.}, 2006.

\bibitem{wachter-mp06}
A.~W{\"a}chter and L.~T. Biegler, ``\href{https://link.springer.com/article/10.1007/s10107-004-0559-y#citeas}{On the implementation of an interior-point filter line-search algorithm for large-scale nonlinear programming},'' \emph{Math. Progr.}, vol. 106, 2006.

\bibitem{HSL}
``{Harwell Subroutine Library, AEA Technology, Harwell, Oxfordshire, England. A catalogue of subroutines},'' \url{http://www.hsl.rl.ac.uk/}.

\bibitem{mastalli22auro}
C.~{Mastalli}, J.~{Marti-Saumell}, W.~{Merkt}, J.~{Sola}, N.~{Mansard}, and S.~{Vijayakumar}, ``\href{https://arxiv.org/pdf/2010.00411.pdf}{A Feasibility-Driven Approach to Control-Limited DDP},'' \emph{Autom. Rob.}, vol.~46, 2022.

\bibitem{dellaert2017factorgraphs}
F.~Dellaert and M.~Kaess, \emph{\href{https://ieeexplorDe.ieee.org/document/8187520}{Factor Graphs for Robot Perception}}.\hskip 1em plus 0.5em minus 0.4em\relax Now Foundations and Trends, 2017.

\bibitem{harley2018contactfactors}
R.~Hartley, J.~Mangelson, L.~Gan, M.~Ghaffari~Jadidi, J.~M. Walls, R.~M. Eustice, and J.~W. Grizzle, ``\href{https://ieeexplore.ieee.org/document/8460748}{Legged Robot State-Estimation Through Combined Forward Kinematic and Preintegrated Contact Factors},'' in \emph{IEEE Int. Conf. Rob. Autom. (ICRA)}, 2018.

\bibitem{agrawal22factorgraph}
V.~Agrawal, S.~Bertrand, R.~Griffin, and F.~Dellaert, ``\href{https://ieeexplore.ieee.org/document/10000099}{Proprioceptive State Estimation of Legged Robots with Kinematic Chain Modeling},'' in \emph{IEEE Int. Conf. Hum. Rob. (ICHR)}, 2022.

\bibitem{bellman54bull}
R.~E. Bellman, \emph{\href{https://www.ams.org/journals/bull/1954-60-06/S0002-9904-1954-09848-8/S0002-9904-1954-09848-8.pdf}{The Theory of Dynamic Programming}}.\hskip 1em plus 0.5em minus 0.4em\relax RAND Corporation, 1954.

\bibitem{kobilarov2015optestddp}
M.~Kobilarov, D.-N. Ta, and F.~Dellaert, ``\href{https://ieeexplore.ieee.org/document/7139279}{Differential dynamic programming for optimal estimation},'' in \emph{IEEE Int. Conf. Rob. Autom. (ICRA)}, 2015.

\bibitem{oshin2022pddp}
A.~Oshin, M.~D. Houghton, M.~J. Acheson, I.~M. Gregory, and E.~A. Theodorou, ``\href{https://www.roboticsproceedings.org/rss18/p046.pdf}{Parameterized Differential Dynamic Programming},'' in \emph{Rob.: Sci. Sys. (RSS)}, 2022.

\bibitem{mastalli-invdynmpc}
C.~Mastalli, S.~P. Chhatoi, T.~Corbéres, S.~Tonneau, and S.~Vijayakumar, ``\href{}{Inverse-Dynamics MPC via Nullspace Resolution},'' \emph{IEEE Trans. Rob. (T-RO)}, vol.~39, 2023.

\bibitem{li2023multshootddp}
H.~Li, W.~Yu, T.~Zhang, and P.~M. Wensing, ``\href{https://ieeexplore.ieee.org/document/10342217}{A Unified Perspective on Multiple Shooting In Differential Dynamic Programming},'' in \emph{IEEE/RSJ Int. Conf. Intell. Rob. Sys. (IROS)}, 2023.

\bibitem{atkeson1986estimation}
C.~G. Atkeson, C.~H. An, and J.~M. Hollerbach, ``\href{https://journals.sagepub.com/doi/abs/10.1177/027836498600500306?journalCode=ijra}{Estimation of Inertial Parameters of Manipulator Loads and Links},'' \emph{The Int. J. of Rob. Res. (IJRR)}, vol.~5, 1986.

\bibitem{carpentier2018analytical}
J.~Carpentier and N.~Mansard, ``\href{https://www.roboticsproceedings.org/rss14/p38.pdf}{Analytical derivatives of rigid body dynamics algorithms},'' in \emph{Rob.: Sci. Sys. (RSS)}, 2018.

\bibitem{singh2022svaderivative}
S.~Singh, R.~P. Russell, and P.~M. Wensing, ``\href{https://ieeexplore.ieee.org/document/9674779}{Efficient Analytical Derivatives of Rigid-Body Dynamics Using Spatial Vector Algebra},'' \emph{IEEE Rob. Autom. Lett. (RA-L)}, vol.~7, 2022.

\bibitem{featherstone2014rigid}
R.~Featherstone, \emph{\href{https://link.springer.com/book/10.1007/978-1-4899-7560-7}{Rigid body dynamics algorithms}}.\hskip 1em plus 0.5em minus 0.4em\relax Springer, 2014.

\bibitem{traversaroidentification}
S.~Traversaro, S.~Brossette, A.~Escande, and F.~Nori, ``\href{https://ieeexplore.ieee.org/document/7759801}{Identification of fully physical consistent inertial parameters using optimization on manifolds},'' in \emph{IEEE/RSJ Int. Conf. Intell. Rob. Sys. (IROS)}, 2016.

\bibitem{rucker2022smooth}
C.~Rucker and P.~M. Wensing, ``\href{https://ieeexplore.ieee.org/document/9690029}{Smooth Parameterization of Rigid-Body Inertia},'' \emph{IEEE Rob. Autom. Lett. (RA-L)}, vol.~7, 2022.

\bibitem{mastalli2022agile}
C.~Mastalli, W.~Merkt, G.~Xin, J.~Shim, M.~Mistry, I.~Havoutis, and S.~Vijayakumar, ``Agile maneuvers in legged robots: a predictive control approach,'' \emph{arXiv preprint arXiv:2203.07554}, 2022.

\bibitem{budhiraja-ichr18}
R.~Budhiraja, J.~Carpentier, C.~Mastalli, and N.~Mansard, ``\href{https://ieeexplore.ieee.org/document/8624925}{Differential Dynamic Programming for Multi-Phase Rigid Contact Dynamics},'' in \emph{IEEE Int. Conf. Hum. Rob. (ICHR)}, 2018.

\bibitem{wensing2017linear}
P.~M. Wensing, S.~Kim, and J.-J.~E. Slotine, ``\href{https://ieeexplore.ieee.org/document/7987066}{Linear Matrix Inequalities for Physically Consistent Inertial Parameter Identification: A Statistical Perspective on the Mass Distribution},'' \emph{IEEE Rob. Autom. Lett. (RA-L)}, vol.~3, 2017.

\bibitem{smartrepot}
S.~Martinez, R.~Griffin, and C.~Mastalli, ``Derivations of multiple shooting optimal estimation algorithm,'' 2024.

\bibitem{GianlucaFrison2013}
G.~Frison, D.~Kouzoupis, J.~B. Jørgensen, and M.~Diehl, ``\href{https://ieeexplore.ieee.org/abstract/document/7798946}{An efficient implementation of partial condensing for Nonlinear Model Predictive Control},'' in \emph{IEEE Conf. on Dec. Cntrl. (CDC)}, 2016.

\bibitem{nocedal-optbook}
J.~{Nocedal} and S.~{Wright}, \emph{\href{https://www.springer.com/gp/book/9780387303031}{Numerical Optimization}}, 2nd~ed.\hskip 1em plus 0.5em minus 0.4em\relax New York, USA: Springer, 2006.

\bibitem{fallon14pronto}
M.~F. Fallón, M.~Antone, N.~Roy, and S.~Teller, ``\href{https://ieeexplore.ieee.org/document/7041346}{Drift-free humanoid state estimation fusing kinematic, inertial and LIDAR sensing},'' in \emph{IEEE Int. Conf. Hum. Rob. (ICHR)}, 2014.

\end{thebibliography}
